\documentclass{article}

\usepackage{microtype}
\usepackage{graphicx}
\usepackage{booktabs}
\usepackage{hyperref}

\usepackage[accepted]{icml2019}
\icmltitlerunning{CompILE: Compositional Imitation Learning and Execution}
\usepackage{wrapfig}

\usepackage{relsize}
\usepackage{tikz}
\tikzset{fontscale/.style = {font=\relsize{#1}}}
\usetikzlibrary{positioning}
\usetikzlibrary{backgrounds}
\usepackage{subcaption}
\usepackage[symbol]{footmisc}

\usepackage{amsmath,amsthm,amssymb}
\usepackage{mathtools}
\usepackage{xcolor}
\usepackage{multirow}
\usepackage{arydshln}
\newcommand{\icmlInternDeepMind}{\textsuperscript{\dag}Work done during an internship at DeepMind. }
\newcommand{\icmlExDeepMind}{\textsuperscript{\#}Work done while employed at DeepMind. }

\DeclareMathOperator*{\argmax}{arg\,max} 
\DeclareMathOperator*{\argmin}{arg\,min} 

\newcommand{\expt}{\mathbb{E}}

\newcommand{\ELBO}{\mathrm{ELBO}}
\newcommand{\cumsum}{\mathrm{cumsum}}

\begin{document}

\twocolumn[
\icmltitle{CompILE: Compositional Imitation Learning and Execution}

\icmlsetsymbol{intern}{\dag}
\icmlsetsymbol{exdm}{\#}

\begin{icmlauthorlist}
\icmlauthor{Thomas Kipf}{am,intern}
\icmlauthor{Yujia Li}{dm}
\icmlauthor{Hanjun Dai}{gt,intern}
\icmlauthor{Vinicius Zambaldi}{dm}
\icmlauthor{Alvaro Sanchez-Gonzalez}{dm}
\icmlauthor{Edward Grefenstette}{fair,exdm}
\icmlauthor{Pushmeet Kohli}{dm}
\icmlauthor{Peter Battaglia}{dm}
\end{icmlauthorlist}

\icmlaffiliation{am}{Informatics Institute, University of Amsterdam, Amsterdam, The Netherlands}
\icmlaffiliation{dm}{DeepMind, London, UK}
\icmlaffiliation{fair}{Facebook AI Research, London, UK}
\icmlaffiliation{gt}{School of Computational Science and Engineering, Georgia Institute of Technology, Atlanta, Georgia, USA}

\icmlcorrespondingauthor{Thomas Kipf}{t.n.kipf@uva.nl}
\icmlkeywords{Machine Learning, ICML}

\vskip 0.3in
]

\printAffiliationsAndNotice{\icmlInternDeepMind\icmlExDeepMind}

\begin{abstract}
We introduce Compositional Imitation Learning and Execution (CompILE): a framework for learning reusable, variable-length segments of hierarchically-structured behavior from demonstration data. CompILE uses a novel unsupervised, fully-differentiable sequence segmentation module to learn latent encodings of sequential data that can be re-composed and executed to perform new tasks. Once trained, our model generalizes to sequences of longer length and from environment instances not seen during training. We evaluate CompILE in a challenging 2D multi-task environment and a continuous control task, and show that it can find correct task boundaries and event encodings in an unsupervised manner. Latent codes and associated behavior policies discovered by CompILE can be used by a hierarchical agent, where the high-level policy selects actions in the latent code space, and the low-level, task-specific policies are simply the learned decoders. We found that our CompILE-based agent could learn given only sparse rewards, where agents without task-specific policies struggle.
\end{abstract}
\section{Introduction}
Discovering compositional structure in sequential data, without supervision, is an important ability in human and machine learning. For example, when a cook prepares a meal, they re-use similar behavioral sub-sequences (e.g., slicing, dicing, chopping) and compose the components hierarchically (e.g., stirring together eggs and milk, pouring the mixture into a hot pan and stirring it to form scrambled eggs).
Humans are adept at inferring event structure by hierarchically segmenting continuous sensory experience \cite{zacks2001perceiving,baldassano2017discovering,radvansky2017event}, which may support building efficient event representations in episodic memory \cite{ezzyat2011constitutes} and constructing abstract plans \cite{richmond2017constructing}. 

\begin{figure}[t!]
  \centering
    \includegraphics[width=0.9\linewidth]{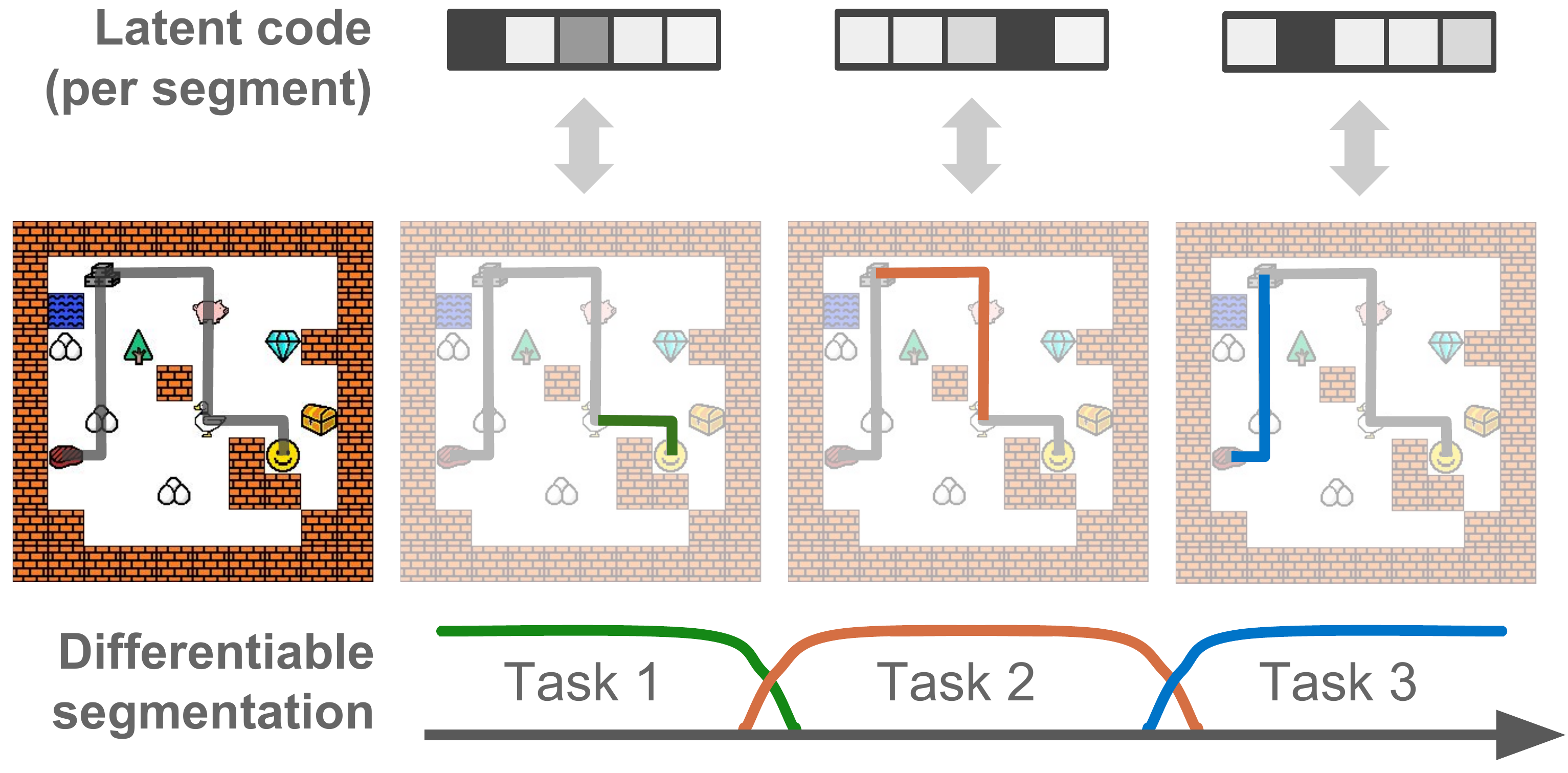}
   \caption{Joint unsupervised learning of task segmentation and encoding in CompILE. CompILE auto-encodes sequential demonstration data by 1) softly breaking an input sequence into segments of variable length, and 2) mapping each such segment into a latent code, which can be executed to reconstruct the input sequence. At test time, the latent code can be re-composed to produce novel behavior.\label{fig:intro_figure}}
\end{figure}

An important benefit of compositional sub-sequence representations is combinatorial generalization to never-before-seen conjunctions \cite{davidson1984compositionality,denil2017programmable}. Behavioral sub-components can also be used as high-level actions in hierarchical decision-making, offering improved credit assignment and efficient planning. 
To reap these benefits in machines, however, the event structure and composable representations must be discovered in an unsupervised manner, as sub-sequence labels are rarely available.

In this work, we focus on the problem of jointly learning to segment, explain, and imitate agent behavior (from demonstrations) via an unsupervised auto-encoding objective. The encoder learns to jointly infer event boundaries and high-level abstractions (latent encodings) of activity within each event segment, while the task of the decoder is to reconstruct or imitate the original behavior by executing the inferred sequence of latent codes. 

We introduce a \emph{fully differentiable}, unsupervised segmentation model that we term CompILE (Compositional Imitation Learning and Execution) that addresses the segmentation problem by predicting soft \textit{segment masks}. During training, the model makes multiple passes over the input sequence, explaining one segment of activity at a time. Segments explained by earlier passes are softly masked out and thereby ignored by the model. 
Our approach to masking is related to soft self-attention  \cite{parikh2016decomposable,vaswani2017attention}, where each mask predicted by our model is localized in time (see Figure \ref{fig:intro_figure} for an example). At test time, these soft masks can be replaced with discrete, consecutive masks that mark the beginning and end of a segment. This allows us to process sequences of arbitrary length by 1) identifying the next segment, 2) explaining this segment with a latent variable, and 3) cutting/removing this segment from the sequence and continue the process on the remainder of the input.

Formally, our model takes the form of a conditional variational auto-encoder (VAE) \cite{kingma2013auto,rezende2014stochastic,sohn2015learning}. We introduce a method for modeling segment boundaries as softly relaxed discrete latent variables \cite{jang2016categorical,maddison2016concrete} which allows for efficient, low-variance training.

We demonstrate the efficacy of our approach in a multi-task, multiple instruction-following domain similar to \citet{oh2017zero} and a continuous control environment. Our model can reliably discover event boundaries and find effective event (sub-task) encodings. In a number of experiments, we found that CompILE generalizes to unseen environment configurations and to task sequences which were longer than those seen during training.

Once trained, the latent codes and associated behavior discovered by CompILE can be reused and recomposed to solve new, unseen tasks. We demonstrate this ability in a set of experiments using a hierarchical agent, with a \emph{meta controller} that learns to operate over discovered policies and associated latent codes to solve difficult sparse reward tasks, where non-hierarchical, non-compositional baselines struggle to learn.

\section{Model overview}
We consider the task of auto-encoding sequential data by 1) breaking an input sequence into disjoint segments of variable length, and 2) mapping each segment individually into some higher-level code, from which the input sequence can be reconstructed.

More specifically, we focus on modeling state-action trajectories of the form $\rho = ((s_1, a_1), (s_2, a_2), ..., (s_T, a_T))$ with states $s_t\in\mathcal{S}$ and actions $a_t\in\mathcal{A}$ for time steps $t = 1, ..., T$, e.g.~obtained from a dataset $\mathcal{D} = \{\rho_1, \rho_2, ..., \rho_N\}$ of $N$ expert demonstrations of variable length for a set of tasks.

\subsection{Behavioral cloning}
Our basic setup follows that of behavioral cloning (BC), i.e., we want to find an imitation policy $\pi_\theta$, parameterized by $\theta$, by solving the following optimization problem:
\begin{equation}
\theta^* = \argmax_\theta \expt_{\rho\in\mathcal{D}}\left[ p_\theta(a_{1:T}|s_{1:T})\right].
\end{equation}
In BC we have $p_\theta(a_{1:T}|s_{1:T}) = \prod_{t=1:T} \pi_\theta(a_t|s_t)$,
where $\pi_\theta(a|s)$ denotes the probability of taking action $a$ in state $s$ under the imitation policy $\pi_\theta$.

\subsection{Sub-task identification and imitation} 
Differently from the default BC setup, our model breaks trajectories $\rho$ into $M$ disjoint segments $(c_1, c_2, ..., c_M)$:
\begin{equation}
c_i = ((s_{b_{i'}}, a_{b_{i'}}), (s_{b_{i'}+1}, a_{b_{i'}+1}), ..., (s_{b_{i}-1}, a_{b_{i}-1})),
\end{equation}
where $M$ is a hyperparameter, and $i'=i-1$. Here, $b_i\in[1, T+1]$ are discrete (latent) boundary indicator variables with $b_0=1$, $b_M=T+1$, and $b_i\geq b_{i'}$. We allow segments $c_i$ to be empty if $b_{i}=b_{i'}$. We model each part independently with a sub-task policy  $\pi_\theta(a|s,z)$, where $z$ is a latent variable summarizing the segment. Framing BC as a joint segmentation and auto-encoding problem allows us to obtain imitation policies that are specific to different inferred sub-tasks, and which can be re-combined for easier generalization to new settings. Each sub-task policy is responsible for explaining a variable-length segment of the demonstration trajectory.

We take the segment (sub-task) encoding $z$ to be discrete in the following, but we note that other choices are possible and require only minor modifications to our framework. The probability of an action sequence $a_{1:T}$ given a sequence of states $s_{1:T}$ then takes the following form\footnote{We again use the shorthand notation $i'=i-1$ for clarity.}:
\begin{align}
\label{eq:generative-model}
&p_\theta(a_{1:T}|s_{1:T}) = \\
&\sum_{b_{1:M}}\sum_{z_{1:M}} p_\theta(a_{1:T}|s_{1:T},b_{1:M},z_{1:M}) p(b_{1:M},z_{1:M}) =\nonumber\\
&\sum_{\substack{b_{1:M}\\z_{1:M}}}\prod_{i=1:M} p_\theta(a_{b_{i'}:b_{i}-1}|s_{b_{i'}:b_{i}-1},z_i) p(b_{i}|b_{i'})p(z_i)  = \nonumber\\[-1em]
& \sum_{\substack{b_{1:M}\\z_{1:M}}}\prod_{i=1:M}\left[\quad\,\,\prod_{\mathclap{j=b_{i'}:b_{i}-1}}\,\,\pi_\theta(a_j|s_j,z_i)\right] p(b_{i}|b_{i'})p(z_i) ,\nonumber
\end{align}
where the double summation marginalizes over all allowed configurations of the discrete latent variables $z_{1:M}$ and $b_{1:M}$. We omit $p(b_0)$ since we set $b_0=1$. Note that our framework supports both discrete and continuous latent variables $z_{1:M}$---for the latter case, the summation sign in Eq.\eqref{eq:generative-model} is replaced with an integral. Our (conditional) generative model $p_\theta(a_{1:T}|s_{1:T},b_{1:M},z_{1:M})$ factorizes across time steps if we choose a non-recurrent policy $\pi_\theta(a|s,z)$. Using recurrent policies is necessary, e.g., for partially observable environments and is left for future work.

For simplicity, we assume independent priors over $b$ and $z$ as follows: $p(b_i,z_i|b_{1:i'},z_{1:i'}):=p(b_i|b_{i'})p(z_i)$. If more complex dependencies are present in the data, this assumption can be replaced with some mechanism for implementing conditional probabilities between segments. We choose a uniform categorical prior $p(z_i)$ and the following empirical categorical prior for the boundary latent variables:
\begin{equation}
p(b_i|b_{i'})\propto \mathrm{Poisson}(b_i - b_{i'}, \lambda) = e^{-\lambda}\frac{\lambda^{b_i - b_{i'}}}{(b_i - b_{i'})!},
\end{equation}
proportional to a Poisson distribution with rate $\lambda$, but truncated to the interval $[b_{i'}, T+1]$ and renormalized, as we are dealing with sequences of finite length. This prior encourages segments to be close to $\lambda$ in length and helps avoid two failure modes: 1) collapse of segments to unit length, and 2) a single segment covering the full sequence length.

\subsubsection{Recognition model}
Following the standard VAE \cite{kingma2013auto,rezende2014stochastic} framework, we introduce a recognition model $q_\phi(b_{1:M},z_{1:M}|a_{1:T},s_{1:T})$ that allows us to infer a task decomposition via boundary variables $b_{1:M}$ and task encodings $z_{1:M}$ for a given trajectory $\rho$. We would like our recognition model to be able to generalize to new compositions of the underlying latent code. We can encourage this by dropping the dependence of $q_\phi$ on any time steps before the previous boundary position. In practice, this means that once a segment (sub-task) has been identified and explained by a latent variable $z$, the corresponding part of the input trajectory will be masked out and the recognition model proceeds on the remainder of the trajectory, until the end is reached. This will further facilitate generalization to sequences of longer length (and with more segments) than those seen during training. 

Formally, we structure the recognition model as follows:
\begin{align}
\label{eq:recognition-model}
q_\phi(b_{1:M},&z_{1:M}|x_{1:T}) =\nonumber\\&\prod_{i=1:M}q_{\phi_z}(z_{i}|x_{b_{i'}:b_i-1})q_{\phi_b}(b_{i}|x_{b_{i'}:T}),
\end{align}
where we have used $x_t=(a_t,s_t)$ and $i'=i-1$ to simplify notation. Expressed in other words, we re-use the same recognition model with shared parameters for each segment while masking out already explained segments. The core modules are the \textit{encoding network} $q_{\phi_z}(z|x)$ and the \textit{boundary prediction network} $q_{\phi_b}(b|x)$, both are modeled as categorical distributions. We use recurrent neural networks (RNN)---specifically, a uni-directional LSTM \cite{hochreiter1997long}---with shared parameters, but with different output heads: one head for predicting the logits $h_{b_i}$ for the boundary latent variable $b_i$ at every time step, and one head for predicting the logits $h_{z_i}$ for the sub-task encoding $z_i$ at the last time step in the current segment $C_i$. 

We use multi-layer perceptrons (MLPs) to implement the output heads:
\begin{align}
\label{eq:recognition-model-rnn}
h_{z_{i}} &= \mathrm{MLP}_z(\mathrm{LSTM}_{b_{i}-1}(\tilde{x}_{b_{i'}:b_{i}-1})),\\\
h_{b_{i}}^t &= \mathrm{MLP}_b(\mathrm{LSTM}_{t}(\tilde{x}_{b_{i'}:T})),
\end{align}
where the MLPs have parameters specific to $b$ or $z$ (i.e., not shared between the output heads). The subscript $t$ on $\mathrm{LSTM}_t$ denotes the time step at which the output is read. Note that $h_{z_i}$ is a $K$-dimensional vector where $K$ is the number of latent categories, whereas $h_{b_i}^t$ is a scalar specific to time step $t$. $\tilde{x}_t$ denotes a learned embedding of the input $x_t$ at time step $t$. In practice, we implement this embedding using a convolutional neural network (CNN), i.e., $\tilde{x}_t=\mathrm{CNN}(x_t)$, with layer normalization \cite{ba2016layer} for pixel-based inputs and using an MLP otherwise. Note that the CNN is only applied to the state, but not on the action component of $x_t$.

\subsubsection{Continuous relaxation}
\begin{figure*}[htp]
  \centering
    \makebox[\textwidth][c]{\includegraphics[width=\textwidth]{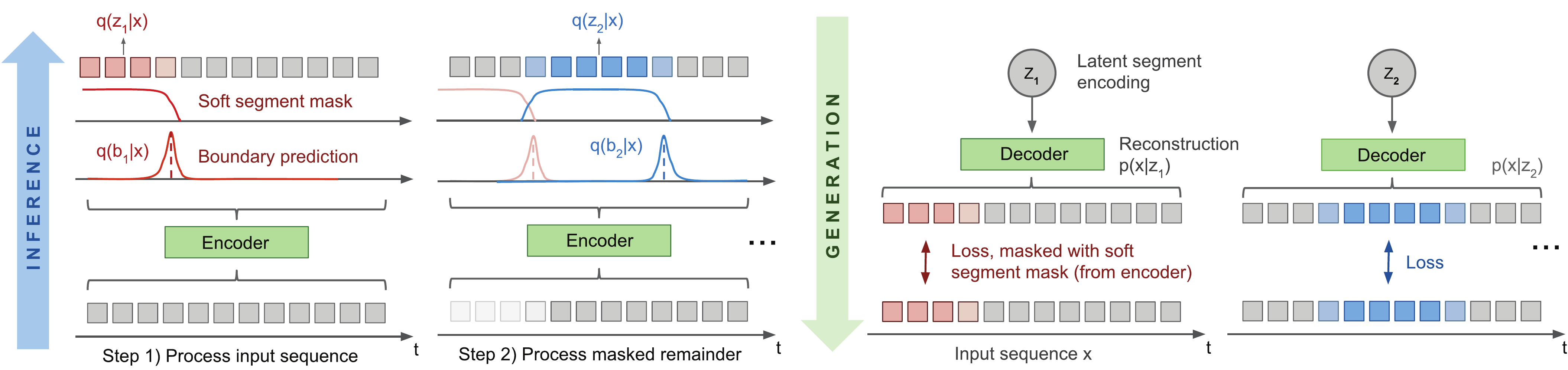}}
  \caption{Differentiable segmentation of an input trajectory $x$ composed of a sequence of sub-tasks. The recognition model (encoder, marked as \textit{inference}) predicts relaxed categorical (Gumbel softmax) boundary distributions $q(b_i|x)$ from which we can obtain soft segment masks $P(t\in C_i)$. Each segment $C_i$ is encoded via $q(z_i|x)$. The generative model $p(x|z_i)$ is executed once for every latent variable $z_i$. The reconstruction loss is masked with $P(t\in C_i)$, so that only the reconstructed part corresponding to the $i$-th segment receives a training signal. For imitation learning, the generative model (decoder, marked as \textit{generation}) takes the form of a policy $\pi_\theta(a_t|s_t,z_i)$. \label{fig:segmentation}}
\end{figure*}

We can jointly train the recognition and the generative model by using the usual ELBO as an objective for learning (see supplementary material). To obtain low-variance gradient estimates for learning, we can use the reparameterization trick for VAEs \cite{kingma2013auto}. Our current model formulation, however, does not allow for reparameterization as both $b$ and $z$ are discrete latent variables. To circumvent this issue, we make use of a continuous relaxation, i.e., we replace the respective categorical distributions with Gumbel softmax / concrete \cite{maddison2016concrete,jang2016categorical} distributions. While this is straightforward for the sub-task latent variables $z$, some extra consideration is required to translate the constraint $b_i \geq b_{i'}$ and the conditioning on trajectory segments of the form $x_{b_{i'}:b_{i}-1}$ to the continuous case. Note that we again summarize pairs of states $s_t$ and actions $a_t$ in a single variable $x_t=(a_t, s_t)$ for ease of notation. The continuous relaxation is only necessary at training time, during testing we can fall back to the discrete version explained in the previous section.

\paragraph{Soft segment masks}
In the relaxed/continuous case at training time we cannot enforce a strict ordering $b_i \geq b_{i'}$ on the boundaries directly as we are now dealing with ``soft'' distributions and don't have access to discrete samples at training time. It is still possible, however, to evaluate segment probabilities of the form $P(t\in C_i)$, i.e., the probability that a certain time step $t$ in the trajectory $\rho$ belongs to the $i$-th segment $C_i=[\max_{0\le j \le i-1} b_j, b_i)$. The lower boundary of the segment is now given by the maximum value of all previous boundary variables, as the ordering $b_i \geq b_{i'}$ is no longer guaranteed to hold. $C_i$ is assumed to be empty if any $b_j\geq b_i$ with $j<i$.  We can evaluate segment probabilities as follows:
\begin{align}
&P(t\in C_i) = P\left(\max_{0\le j \le i-1}b_j\le t< b_{i}\right) = \\
&\left[1-\cumsum(q_{\phi_b}(b_{i}|x), t)\right]\,\,\prod_{\mathclap{j=0:i-1}}\,\,\cumsum(q_{\phi_b}(b_{j}|x), t),\nonumber
\end{align}
where $\cumsum(q_{\phi_b}(b_{j}|x), t)=\sum_{k\le t} q_{\phi_b}(b_{j}=k|x)$ is a shorthand for the \textit{inclusive} cumulative sum of the posterior $q_{\phi_b}(b_{j}|x)$, evaluated at time step $t$, i.e., it is equivalent to the CDF of $q_{\phi_b}(b_{j}|x)$. We further have $\cumsum(q_{\phi_b}(b_{0}|x), t) = 1$ and $\cumsum(q_{\phi_b}(b_{M}|x), t) = 0$. It is easy to verify that $\sum_{i=1:M}P(t\in C_i)=1$ for all $t$. These segment probabilities can be seen as soft segment masks. See Figure \ref{fig:segmentation} for an example.

\paragraph{RNN state masking}
We softly mask out parts of the input sequence explained by earlier segments. Using a soft masking mechanism allows us to find suitable segment boundaries via backpropagation, without the need to perform explicit and potentially expensive/intractable marginalization over latent variables. Specifically, we mask out the \textit{hidden states}\footnote{Including the cell state in the LSTM architecture.} of the encoding and boundary prediction networks' RNNs. Thus, inputs belonging to earlier segments are effectively hidden from the model while still allowing gradients to be passed through. The hidden state mask for the $i$-th segment takes the following form:
\begin{align}
&\mathrm{mask}_i(t) = P\left(t \geq \max_{0\le j\le i-1}b_j\right) = \\
&\prod_{j=0:i-1}P(t\geq b_{j}) = \prod_{j=0:i-1}\cumsum(q_{\phi_b}(b_{j}|x), t),\nonumber
\end{align}
where we set $\mathrm{mask}_1=1$. In other words, it is given by the probability for a given time step to \emph{not} belong to a previous segment. Masking is performed by multiplying the RNN's hidden state with $\mathrm{mask}_i$ (after the RNN update of the current time step). For every segment $i\in[1, M]$ we thus need to run the RNN over the full input sequence, while multiplying the hidden states with a segment-specific mask. Nonetheless, the parameters of the RNN are shared over all segments.

\paragraph{Soft RNN readout}
In addition to softly masking the RNN hidden states in both $q_{\phi_b}(b_i|x)$ and $q_{\phi_z}(z_i|x)$, we mask out illegal boundary positions by setting the respective logits to a large negative value. Specifically, we mask out the first time step (as any boundary placed on the first time step would result in an empty segment) and any time steps corresponding to padding values when training on mini-batches of sequences with different length. We allow boundaries (as they are exclusive) to be placed at time step $T+1$. Further, to obtain $q_{\phi_z}(z_i|x)$ from the $z$-specific output head $h_z^t$---where $t$ denotes the time step at which we are reading from the RNN---we perform the following weighted average:
\begin{align}
\label{eq:output_head_masking}
q_{\phi_z}(z_i|x) = \mathrm{concrete}_\tau\left(\sum_{t=1:T} q_{\phi_b}(b_i=t+1|x) \,h_{z_i}^t\right),
\end{align}
which can be understood as the ``soft'' equivalent of reading the output head $h_z^t$ for the last time step within the corresponding segment. $\mathrm{concrete}_\tau$ is a concrete / Gumbel softmax distribution \cite{jang2016categorical,maddison2016concrete} with temperature $\tau$. Note the necessary shift of the boundary distribution by 1 time step, as $q_{\phi_b}(b_i|x)$ points to the first time step of the \emph{following} segment.

\paragraph{Loss masking}
The reconstruction loss part of the ELBO $\mathcal{L}=-\expt_{q_\phi(b, z|a,s)}[\log p_\theta(a|s,b,z)]$ decomposes into independent loss terms for each segment, i.e., $\mathcal{L}=\sum_{i=1:M}\mathcal{L}_i$, due to the structure of our generative model, Eq.~\eqref{eq:generative-model}. To retain this property in the relaxed/continuous case, we softly mask out irrelevant parts of the action trajectory when evaluating the loss term for a single segment:
\begin{align}
\label{eq:loss_masking}
\mathcal{L}_i = \expt_{q_\phi(b,z|a,s)}[\mathrm{seg}_i\cdot\log p_{\theta}(a|s,z_i)],
\end{align}
where the segment mask for time step $t$ is given by $\mathrm{seg}_i(t)=P(t\in C_i)$, i.e.~the probability of time step $t$ being explained by the $i$-th segment. The operator ``$\cdot$'' denotes element-wise multiplication. In practice, we use a single sample of the (reparameterized) posterior to evaluate Eq.~\eqref{eq:loss_masking}.

\paragraph{Number of segments}
At training time, we need to specify the maximum number of segments $M$ that the model is allowed to use when auto-encoding a particular sequence of length $T$. For efficient mini-batch training, we choose a single, fixed $M$ for all training examples. Providing the correct number of segments can further be utilized as a form of weak supervision.

\paragraph{Complexity}
Evaluating the model components $q_{\phi_b}(b_i|x)$, $q_{\phi_z}(z_i|x)$, and  $p_{\theta}(x|z_i)$ is $\mathcal{O}(T)$ for a single $i={1,...,M}$. The overall forward pass of the CompILE model for a single demonstration trajectory in terms of its length $T$ and the number of segments $M$ is therefore $\mathcal{O}(TM)$.

\section{Related work}
Our framework is closely related to option discovery \cite{niekum2013incremental,kroemer2015towards,fox2017multi,hausman2017multi,krishnan2017ddco,fox2018parametrized}, with the main difference being that our inference algorithm is agnostic to what type of option (sub-task) encoding is used. Our framework allows for inference of continuous, discrete or mixed continuous-discrete latent variables. \citet{fox2017multi} introduce an EM-based inference algorithm for option discovery in settings similar to ours, however limited to discrete latent variables and to inference networks that are independent of the position of task boundaries: in their case without recurrency and only dependent on the current state/action pair. Their framework was later applied to continuous control tasks \cite{krishnan2017ddco} and neural program modeling \cite{fox2018parametrized}.

Option discovery has also been addressed in the context of inverse reinforcement learning (IRL) using generative adversarial networks (GANs) \cite{goodfellow2014generative} to find structured policies that are close to demonstration sequences \cite{hausman2017multi,sharma2018directed}. This approach requires being able to interact with the environment for imitation learning, whereas our model is based on BC and works on offline demonstration data.

Various solutions for supervised sequence segmentation or task decomposition exist which require varying degrees of supervision \cite{graves2012supervised,escorcia2016daps,krishna2017dense,shiarlis2018taco}. In terms of two recent examples, \citet{krishna2017dense} assume fully-annotated event boundaries and event descriptions at training time whereas TACO \cite{shiarlis2018taco} only requires \emph{task sketches} (i.e., supervision on sub-task encodings but not on task boundaries) and solves an alignment problem to find a suitable segmentation. A related recent approach decomposes demonstration sequences into underlying programs \cite{sun2018neural} in a fully-supervised setting, based on a seq2seq \cite{sutskever2014sequence,vinyals2015show} model without explicitly modeling segmentation.

Outside of the area of \emph{learning from demonstration}, hierarchical reinforcement learning \cite{sutton1999between,kulkarni2016hierarchical,bacon2017option,florensa2017stochastic,vezhnevets2017feudal,riemer2018learning} and the options framework \cite{sutton1999between,kulkarni2016hierarchical,bacon2017option,riemer2018learning} similarly deal with learning segmentations and representations of behavior, but in a purely generative way. Learning with task sketches \cite{andreas2016modular} and learning of transition policies \cite{lee2018composing} has also been addressed in this context.

Unsupervised segmentation and encoding of sequential data has also received considerable attention in natural language and speech processing \cite{blei2001topic,goldwater2009bayesian,chan2016latent,wang2017sequence,tang2018subgoal}, and in the analysis of sequential activity data \cite{johnson2016composing,dai2016recurrent}. In concurrent work, \citet{pertsch2019keyin} introduced a differentiable model for keyframe discovery in sequence data, which is related to our setting. Sequence prediction models with adaptive step size \cite{neitz2018adaptive,jayaraman2018time} can provide segment boundaries as well, but do not directly learn a policy or latent encodings.

\section{Experiments}
The goals of this experimental section are as follows: 1) we would like to investigate whether our model is effective at both learning to find task boundaries and task encodings while being able to reconstruct and imitate unseen behavior, 2) test whether our modular approach to task decomposition allows our model to generalize to longer sequences with more sub-tasks at test time, and 3) investigate whether an agent can learn to control the discovered sub-task policies to quickly learn new tasks in sparse reward settings.

\subsection{Multi-task environments}
We evaluate our model in a fully-observable 2D multi-task grid world, similar to the one introduced in \citet{oh2017zero} and a continuous control task, where a reacher arm has to reach certain target locations. An example instance for each environment is shown in Figure \ref{fig:envirnoment}. See supplementary material for additional implementation and evaluation details.

\begin{figure}[t]
  \centering
  \begin{subfigure}[b]{0.45\linewidth}
        \centering
        \includegraphics[width=0.85\linewidth]{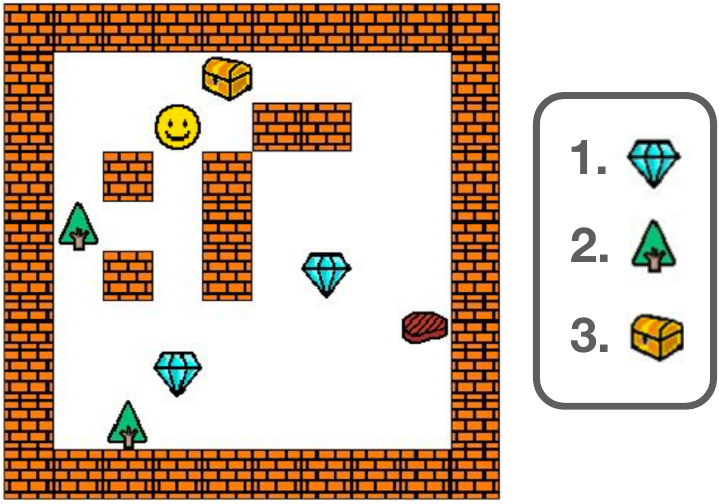}
    \end{subfigure}
      \begin{subfigure}[b]{0.435\linewidth}
        \centering
        \includegraphics[width=0.85\linewidth]{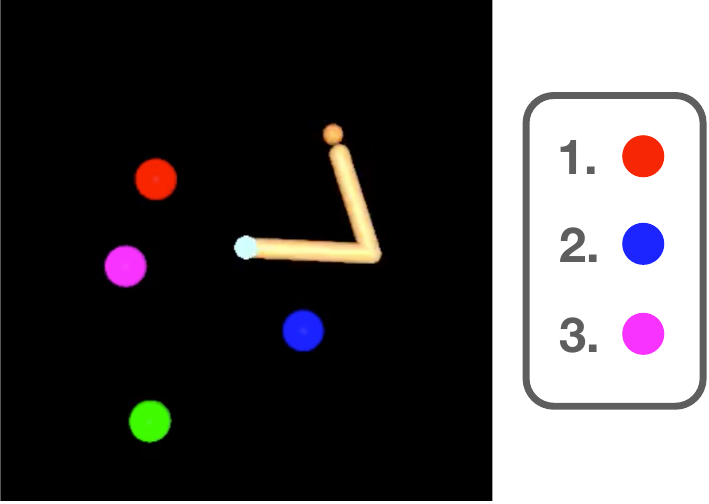}
    \end{subfigure}
  \caption{Example instances of multi-task, instruction-following environments used in our experiments. \textit{Left}: Grid world with walls. An agent has to pick up or visit certain objects. \textit{Right}: Continuous control reacher task with multiple targets. The tip of the reacher arm has to touch multiple colored spheres in a pre-specified order.\label{fig:envirnoment}}
  \vspace{-10pt}
\end{figure}
\paragraph{Grid world}  The environment is a 10x10 grid world with a single agent, impassable walls, and multiple objects scattered throughout the scene. We generate scenes with 6 objects selected uniformly at random from 10 different object types (excl.~walls and player) jointly with task lists of 3-5 \emph{visit} and \emph{pick up} tasks. A single visit task can be solved by moving the agent to the location of an object of the correct type. For example, if the instruction is \emph{visit tree}, the task is completed if any tree in the scene is visited. Similarly, a pick up task can be solved by picking up an object of the correct type (moving to a field adjacent to the object and executing a directional pick up action, e.g.~\emph{pick up north}). We generate a demonstration trajectory for each environment instance and task list by running a shortest path algorithm on the 2D environment grid (while marking walls as impassable).

\paragraph{Continuous control} In this environment, a two-link planar \textit{reacher} arm has to be controlled to reach towards pre-specified target locations. The environment is an adaptation of the single-target reacher task from the DeepMind Control Suite \cite{tassa2018deepmind}. We simultaneously place up to 6 targets drawn without replacement from 10 different target types (spheres of different color) in a single environment instance, distributed uniformly at random within reach of the reacher arm.
The number of targets in an environment is drawn uniformly in range [number of tasks, 6]. For each such instance, we generate a task list by selecting 3-5 of the target object types in the environment. The current target is marked as reached and removed from the scene if the end effector---a small sphere at the tip of the reacher arm---touches the target sphere.  The observations to the agent are the positions of the all targets, and the position of the reacher arm.  We generate demonstration trajectories using a hand-coded control policy, which opens or closes the arm based on the distance of the target to the center, and rotates the shoulder based on the direction to the target.

\subsection{Imitation learning}
\label{sec:imitation-learning}

\begin{figure*}[t!]
    \quad\quad\quad\quad\begin{subfigure}[b]{0.35\textwidth}
        \centering
        \includegraphics[width=\textwidth]{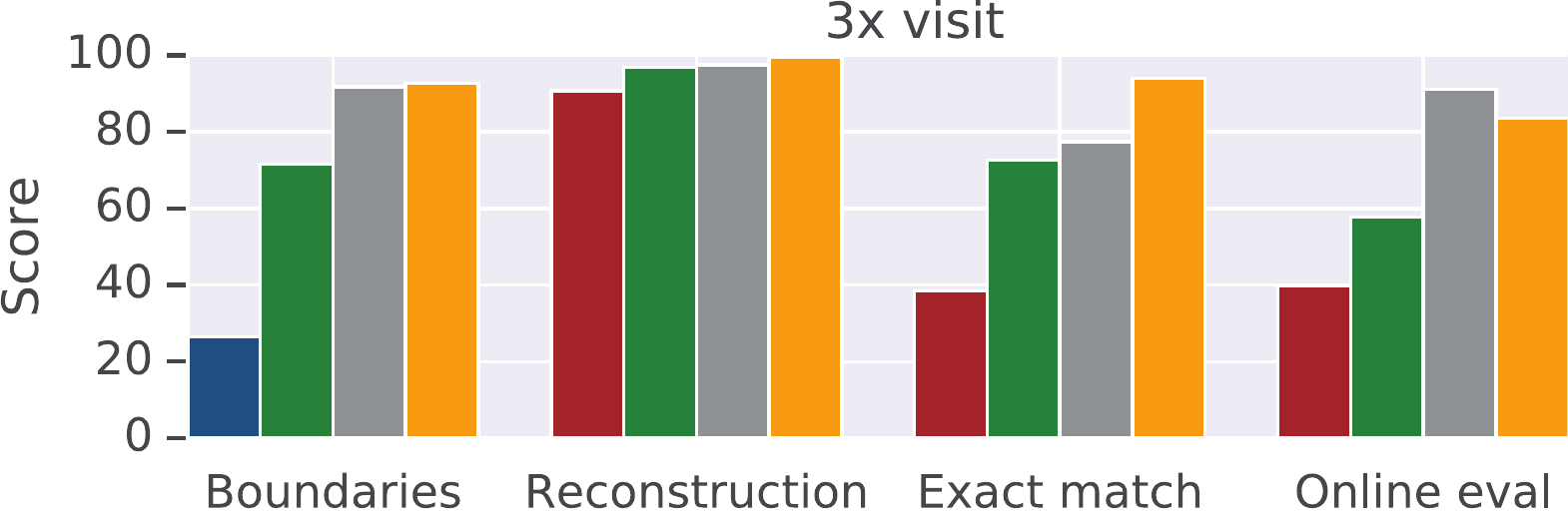}
    \end{subfigure}%
    ~\quad
    \begin{subfigure}[b]{0.47\textwidth}
        \centering
        \includegraphics[width=\textwidth]{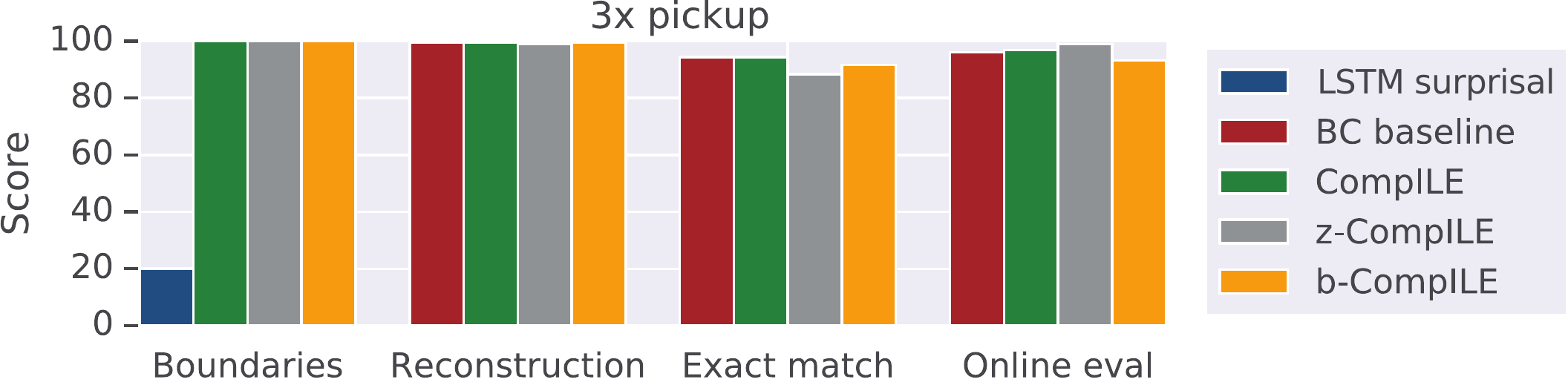}
    \end{subfigure}%
    \\\vspace{0.01em}
    \,\,\,\,\quad\quad\quad\begin{subfigure}[b]{0.35\textwidth}
        \centering\vspace{0.5em}
        \includegraphics[width=\textwidth]{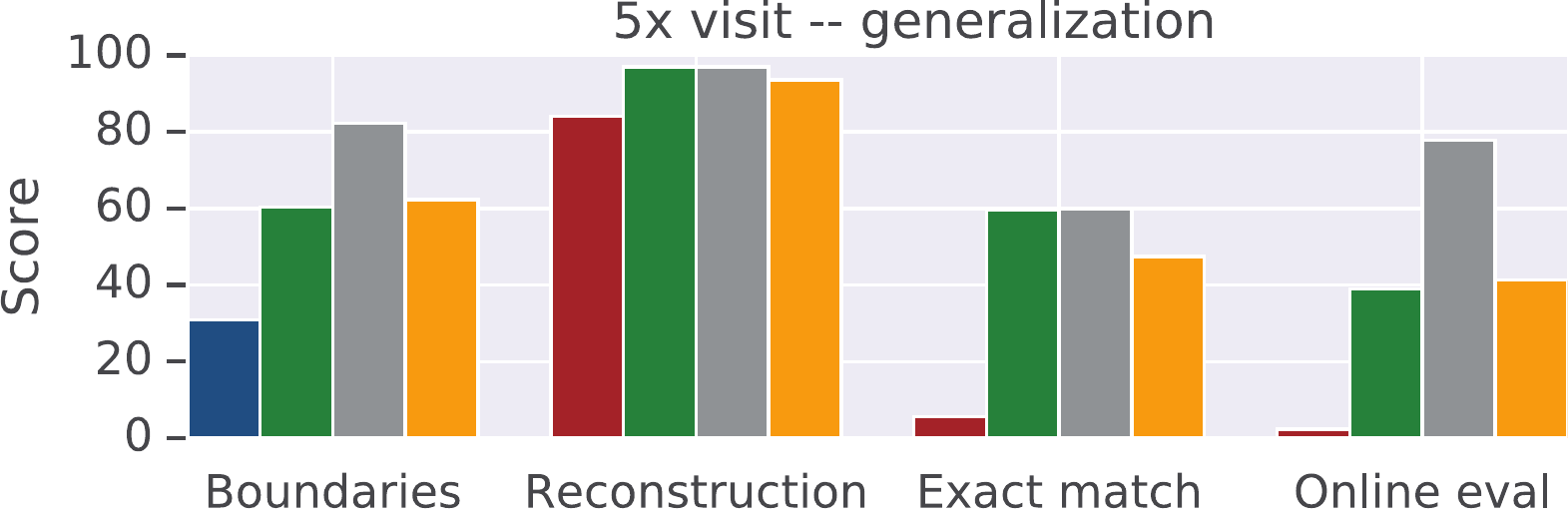}
    \end{subfigure}%
    ~\quad
    \begin{subfigure}[b]{0.35\textwidth}
        \centering\vspace{0.5em}
        \includegraphics[width=\textwidth]{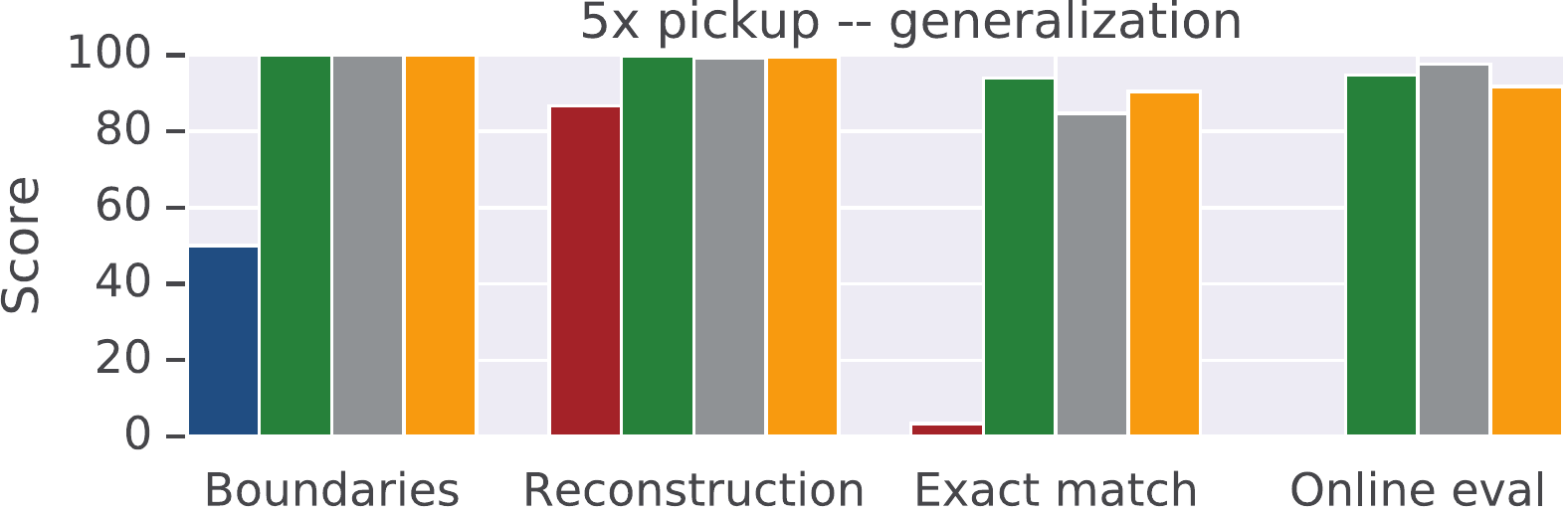}
    \end{subfigure}%
    
    {\centering
    \caption{Imitation learning results in grid world domain. We report accuracy of segmentation boundary recovery, reconstruction accuracy (average over sequence vs.~percentage of exact full-sequence matches) and \emph{online evaluation}: average reward obtained when deploying the generative model (with termination policy) using the inferred latent code from the demonstration sequence in the environment, without re-training. See main text for additional details. \label{fig:imitation_results}}}
\end{figure*}

In this set of experiments, we fit our CompILE model to demonstration trajectories generated for random instances of the multi-task environments (incl.~randomly generated task lists). We train our model with discrete latent variables (as the target types are discrete) on demonstration trajectories with three consecutive tasks, either 3x visit instructions or 3x pick up instructions in the grid world, and 3x reaching instructions in the continuous control environment. Training is carried out on a single GPU with a fixed learning rate of $10^{-4}$ using the Adam \cite{kingma2014adam} optimizer, with a batch size of 256 and for a total of 50k training iterations (500k for reacher task). We further train a causal termination policy that shares the same architecture as the encoder of CompILE to mimic the boundary prediction module in an online setting, i.e., without \textit{seeing the future}.

We evaluate our model on 1024 newly generated instances of the environment. We again generate demonstration trajectories with random task lists of either 3 consecutive tasks (same number as during training) or 5 consecutive tasks, to test for generalization to longer sequences, and we evaluate both boundary prediction performance and accuracy of action sequence reconstruction from the inferred latent code. We provide weak supervision by setting the number of segments to $M=3$ and $M=5$, respectively. We find that results slightly degrade with non-optimal choice of $M$ (see additional experiments in the supplementary material).

\paragraph{Baselines} We compare against two baselines that are based on behavioral cloning (BC): an autoregressive baseline for evaluating segmentation performance, termed \emph{LSTM surprisal}, where we find segment boundaries by thresholding the state-conditional likelihood of an action. In the grid world domain, we further compare against a VAE-based \emph{BC baseline} that corresponds to a variant of our model without inferred task boundaries, i.e.~with only a single segment. This baseline allows us to evaluate task reconstruction performance from an expert trajectory that is encoded in a \emph{single} latent variable. We choose a 32-dim.~Gaussian latent variable $z$ (i.e., with significantly higher capacity) and a unit-variance, zero-mean Gaussian prior for this baseline. We further show results for two model variants: z- and b-CompILE, where we provide supervision on the latent variables $z$ or $b$ during training. z-CompILE is comparable to TACO \cite{shiarlis2018taco}, where task sketches ($z$ in our case) are provided both during training and testing (we only provide $z$ during training), whereas b-CompILE is related to imitation learning of annotated, individual tasks.

\paragraph{Grid world results} Results for the grid world tasks are summarized in Figure \ref{fig:imitation_results}. For the pick up task, we see that our model reliably finds the correct boundary positions, i.e., it discovers the correct segments of behavior both in the 3-task setting (same as training) and in the longer 5-task setting. Reconstructions from the latent code sequence are almost perfect and only degrade slightly in the generalization setting to longer sequences, whereas the BC baseline without segmentation mechanism completely fails to generalize to longer sequences (see \emph{exact match} score). In the visit task setting, ground truth boundary positions can be ambiguous (the agent can walk over an object unintentionally on its way somewhere else) which is reflected in the sometimes lower online evaluation score, as the termination policy can be sensitive to ambiguous termination conditions (e.g., unintentionally walked-over objects). Nonetheless, CompILE is often able to generalize to longer sequences whereas the baseline model without task segmentation consistently fails. In both tasks, our model beats a surprisal-driven segmentation baseline by a large margin. 

\paragraph{Continuous control results}  Results for unsupervised segmentation boundary recovery for the reacher task are summarized in Table \ref{tab:cont_control_results}. We find that CompILE can (almost) perfectly recover segmentation boundaries when trained with partial supervision on $z$ (z-CompILE), matching the performance of b-CompILE that receives supervision on boundary position. Note that different from TACO \cite{shiarlis2018taco}, no supervision is provided at test time. The fully unsupervised model (CompILE) outperforms an auto-regressive baseline (LSTM surprisal) by a large margin, but often does not recover the exact segmentation that generated the trajectory. The F1 score with tolerance for misplaced boundaries by 1 time step (tol=1) shows that in some cases the error can be explained by a minor prediction offset. We omit reconstruction performance results in the continuous domain, as a fair evaluation would require addressing the \emph{covariate shift} problem in BC to allow the policy to recover from small errors, e.g., using a technique such as DART \cite{laskey2017dart} to inject noise in the training process. We leave this for future work.

\begin{table}[t!]
\centering
\resizebox{0.95\columnwidth}{!}{%
\begin{tabular}{rccc}
\toprule
\textbf{Model} & \textbf{Accuracy} & \textbf{F1 (tol=0)} & \textbf{F1 (tol=1)}  \\
\midrule
\multicolumn{4}{c}{3 tasks} \\
LSTM surprisal & $ 24.8 \pm 0.6 $ & $ 39.0 \pm 0.3 $ & $ 47.1 \pm 0.4 $  \\
CompILE & $ 62.0 \pm 4.5 $ & $ 74.3 \pm 3.3 $ & $ 78.9 \pm 2.5 $ \\
z-CompILE & $ 99.5 \pm 0.2 $ & $ 99.7 \pm 0.2 $ & $ 99.8 \pm 0.1 $ \\
b-CompILE & $ 99.8 \pm 0.1 $ & $ 99.9 \pm 0.1 $ & $ 100 \pm 0.0 $ \\
\midrule
\multicolumn{4}{c}{5 tasks -- generalization} \\
LSTM surprisal & $ 21.6 \pm 0.5 $ & $ 44.9 \pm 0.5 $ & $ 54.4 \pm 0.5 $ \\
CompILE & $ 41.7 \pm 8.0 $ & $ 69.3 \pm 4.7 $ & $ 74.0 \pm 4.6 $ \\
z-CompILE & $ 98.4 \pm 0.5 $ & $ 99.3 \pm 0.2 $ & $ 99.8 \pm 0.1 $ \\
b-CompILE & $ 98.8 \pm 0.3 $ & $ 99.5 \pm 0.1 $ & $ 99.8 \pm 0.1 $ \\
\bottomrule
\end{tabular}}
\caption{Segmentation results in continuous control domain. We report accuracy (mean and standard deviation over 5 runs) of exact segmentation boundary recovery and two F1 scores (in $\%$), which measure the harmonic mean between precision and recall for boundary prediction, with (tol=1) and without (tol=0) tolerance for boundaries that are misplaced by 1 time step in either direction.\label{tab:cont_control_results}}
\vspace{-1em}
\end{table}

\subsection{Hierarchical reinforcement learning}
In this set of experiments, we pre-train a CompILE model under the same setting as in Section \ref{sec:imitation-learning} in the grid world environment and only keep the discovered sub-task policies and the termination policy. We provide these policies to a hierarchical agent that can either call a low-level action (such as move or pick up) directly in the environment, or call a \emph{meta action}, that executes a particular sub-task policy incl.~termination policy, until a termination criterion is met (termination probability larger than 0.5 or end of episode).

\begin{figure*}[t!]
    \centering
    \begin{subfigure}[b]{0.24\textwidth}
        \centering
        \includegraphics[width=\textwidth,trim={0.5cm 0.5cm 0.5cm 0.5cm},clip]{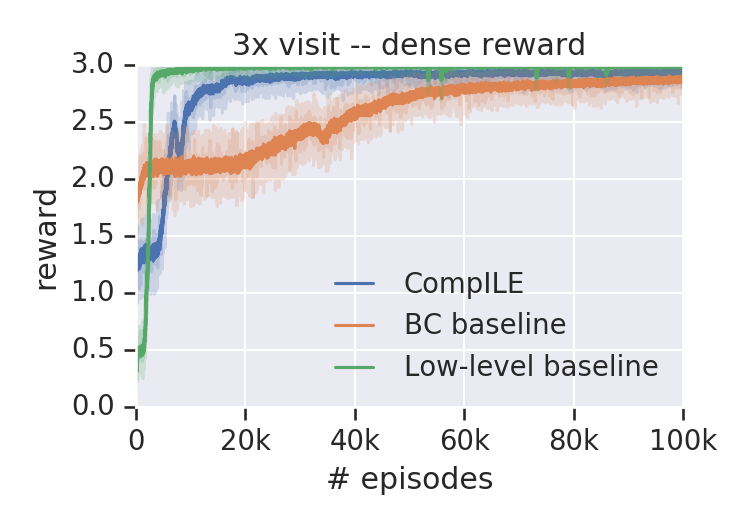}
    \end{subfigure}%
    ~ 
    \begin{subfigure}[b]{0.24\textwidth}
        \centering
        \includegraphics[width=\textwidth,trim={0.5cm 0.5cm 0.5cm 0.5cm},clip]{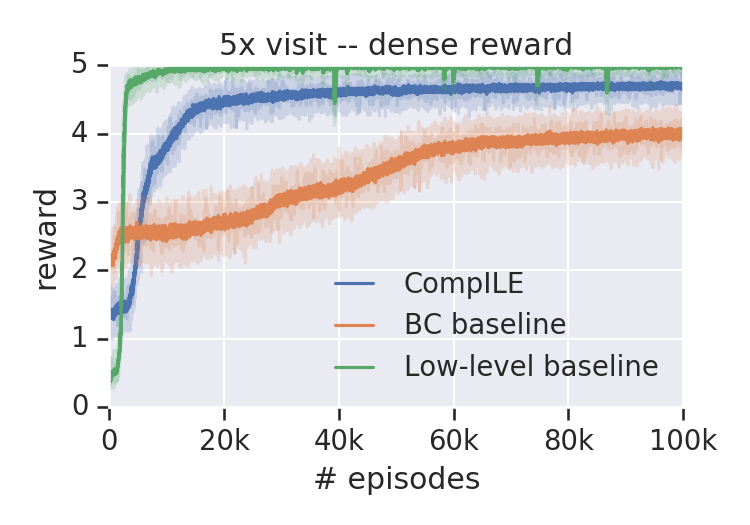}
    \end{subfigure}%
    ~ 
    \begin{subfigure}[b]{0.24\textwidth}
        \centering
        \includegraphics[width=\textwidth,trim={0.5cm 0.5cm 0.5cm 0.5cm},clip]{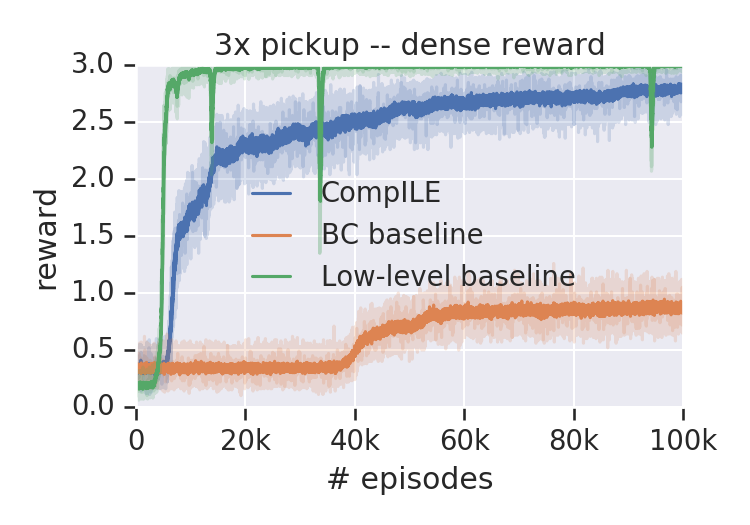}
    \end{subfigure}%
    ~ 
    \begin{subfigure}[b]{0.24\textwidth}
        \centering
        \includegraphics[width=\textwidth,trim={0.5cm 0.5cm 0.5cm 0.5cm},clip]{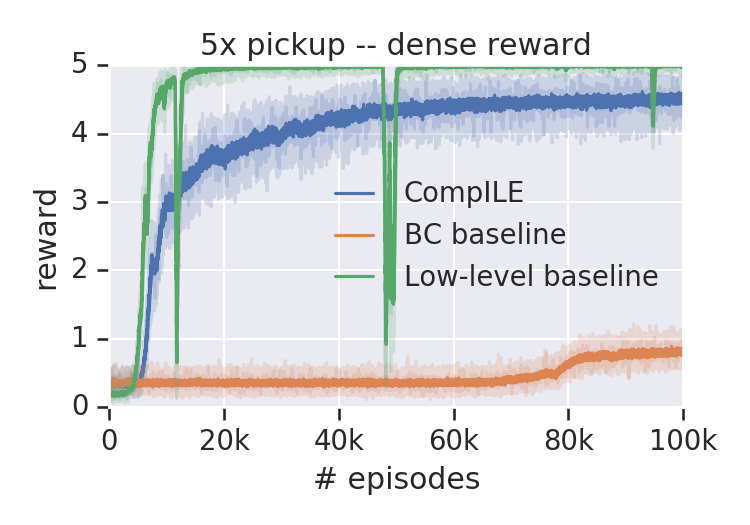}
    \end{subfigure}
    \\\vspace{0.5em}
    \begin{subfigure}[b]{0.24\textwidth}
        \centering
        \includegraphics[width=\textwidth,trim={0.5cm 0.5cm 0.5cm 0.5cm},clip]{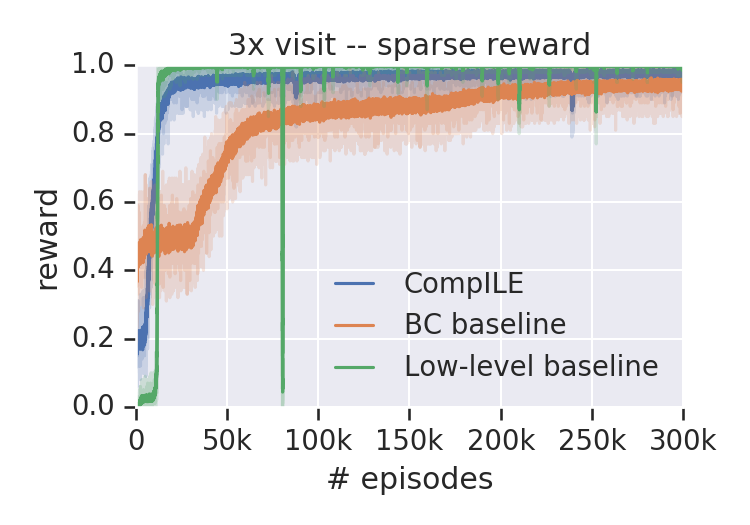}
    \end{subfigure}%
    ~ 
    \begin{subfigure}[b]{0.24\textwidth}
        \centering
        \includegraphics[width=\textwidth,trim={0.5cm 0.5cm 0.5cm 0.5cm},clip]{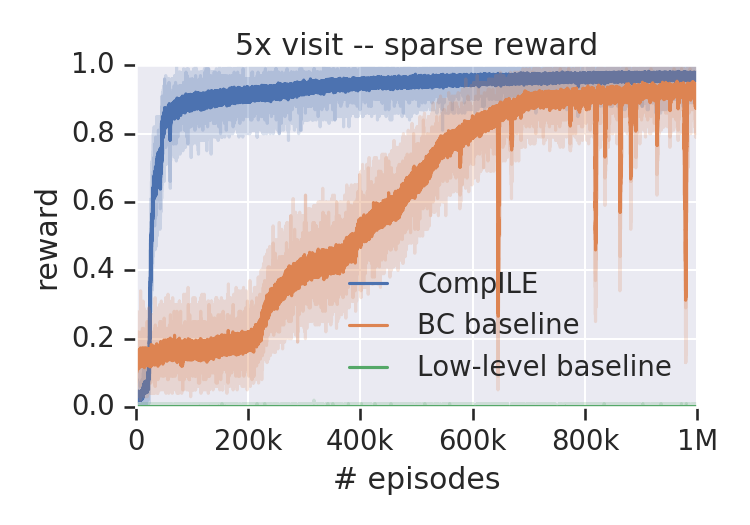}
    \end{subfigure}%
    ~ 
    \begin{subfigure}[b]{0.24\textwidth}
        \centering
        \includegraphics[width=\textwidth,trim={0.5cm 0.5cm 0.5cm 0.5cm},clip]{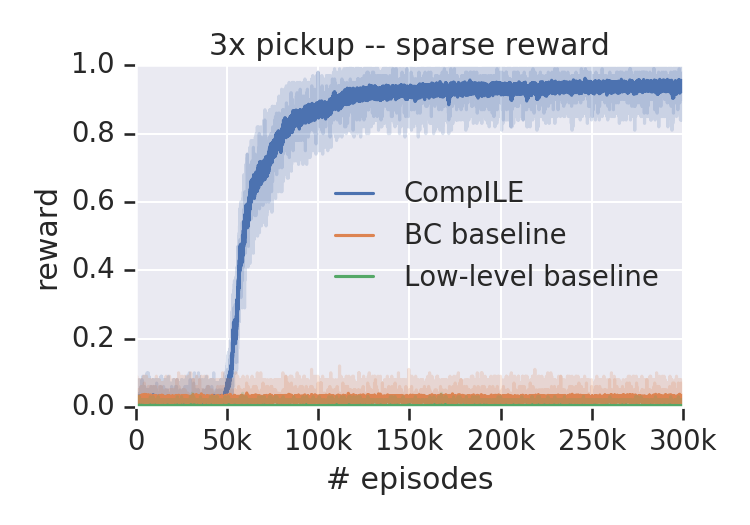}
    \end{subfigure}%
    ~ 
    \begin{subfigure}[b]{0.24\textwidth}
        \centering
        \includegraphics[width=\textwidth,trim={0.5cm 0.5cm 0.5cm 0.5cm},clip]{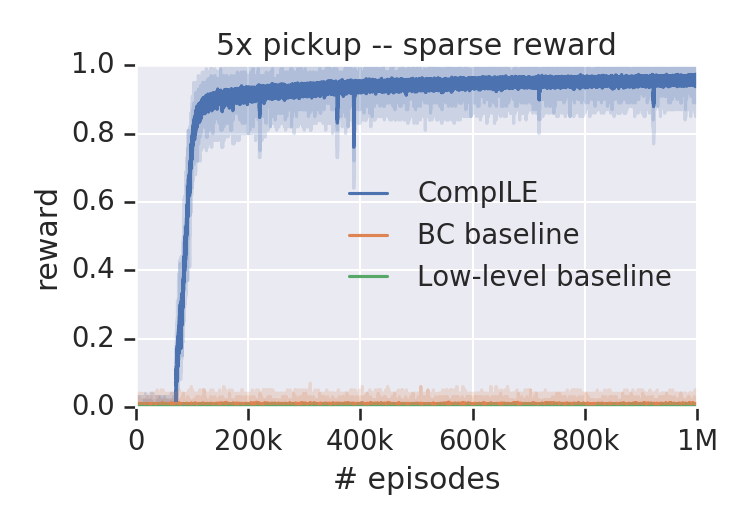}
    \end{subfigure}
    \caption{Learning curves for agents trained in the multi-task grid world environment for a single representative seed. We found that the qualitative behavior was consistent across seeds. Original learning curve (reward at every episode) plotted as shaded line; overlaid with solid line using exponential smoothing for easier visibility. BC denotes a VAE-based behavioral cloning baseline that was exposed to the same number of task demonstrations as our CompILE model. The low-level baseline is an agent without internal hierarchy. The CompILE-based hierarchical agent benefits from significantly improved exploration and is the only agent that succeeds at all sparse reward tasks. Best viewed in color.\label{fig:rl-results}}
\end{figure*}

We generate tasks and environments at random as in the imitation learning setting, but deploy agents in the environment where they either receive a reward of 1 for every completed sub-task (\emph{dense reward} setting) or a single reward of 1 at the end of the episode if all tasks are completed and no termination criterion (e.g., wrong object was picked up, or reached maximum number of 50 steps) was met (\emph{sparse reward} setting). The sparse reward setting poses a very challenging exploration problem: the agent only receives a learning signal if it has completed all tasks from the task list in the correct order, without mistakes (i.e., without picking up a wrong object which could render the episode unsolvable). We compare against a low-level baseline agent that only has access to low-level actions and a VAE-based, pre-trained BC baseline that receives the same pre-training as our CompILE agent, but does not learn a task segmentation (it also has access to low-level actions). All agents use the same CNN-based architecture (see supplementary material for details) and are trained using the distributed policy-gradient algorithm IMPALA \cite{espeholt2018impala}. Results are summarized in Figure \ref{fig:rl-results}.

The hierarchical agent with sub-task policies from the CompILE model achieves consistent results across all settings and generalizes well to the 5 task setup, even though it has only seen demonstrations of 3 tasks during pre-training. It is the only agent that learns to solve the pick up task setting with sparse reward. The visit task is significantly easier to solve as the episode does not end if a wrong object is visited. Nonetheless, the low-level baseline (without pre-training) fails to learn under the sparse reward setting for all but the \emph{3x visit} task. Only if reward for every individual sub-task is provided, the low-level baseline learns to solve the task in the fewest number of episodes.

\subsection{Limitations and future work}
As our training procedure is completely unsupervised, the model is free to choose any type of semantics for its latent code. For example, in the grid world environment we found that the model learns a location-specific latent code (with only a small degree of object specificity), whereas the ground truth task list is specific to object type. See supplementary material for an example. It remains to be seen to what degree the latent code can be grounded in a particular manner with only weak supervision, e.g.~in a semi-supervised setting or using pairs of demonstrations with the same underlying task list.
Furthermore, we have currently only explored fully-observable, Markovian settings. An extension to partially-observable environments will likely introduce further challenges, as the generative model will require some form of recurrency or memory, and the model might learn to ignore the latent code altogether.

\section{Conclusions}
Here we introduced CompILE, a model for discovering and imitating sub-components of behavior in sequential demonstration data. Our results showed that CompILE can successfully discover sub-tasks and their boundaries in an imitation learning setting, and the latent sub-task encodings can then be used as sub-policies in a hierarchical RL agent to solve challenging sparse reward tasks. While here we explored imitation learning, where inputs to the model are state-action sequences, in principle our method can be applied to any sequential data, and an interesting future direction is to apply our differentiable segmentation and auto-encoding mechanism to other data domains. Future work will investigate extensions for partially-observable environments, its applicability as an episodic memory module, and a hierarchical extension for abstract, high-level planning.

\subsection*{Acknowledgements}
We would like to thank Junhyuk Oh, Nicolas Heess, Ziyu Wang, Razvan Pascanu, Caglar Gulcehre, Klaus Greff, Neil Rabinowitz, Andrea Tacchetti, Daniel Mankowitz, Chris Burgess, Irina Higgins, Murray Shanahan, Matthew Willson, Matt Botvinick, and Jessica Hamrick for helpful discussions.

\bibliographystyle{icml2019}
\bibliography{references}

\appendix 
\section{CompILE model details}
\subsection{Encoder architecture}
\paragraph{Grid world} Both the recognition model and the generative model (i.e., the sub-task policies) use a two-layer CNN with $3\times3$ filters and 64 feature maps in each layer, followed by a ReLU activation each. 
We flatten the output representation into a vector and pass it through another trainable linear layer, without activation function. Only for the recognition model, we further concatenate a linear (trainable) embedding of the action ID to this representation. In all cases, we pass the output through a LayerNorm \cite{ba2016layer} layer before it is passed on to other parts of the model, e.g.~the RNN in the recognition model or the sub-task policy MLP in the generative model. 

The LSTM state of the recognition model is reset to 0 between trajectories (and after each pass over the trajectory, i.e., for each segment).
\paragraph{Continuous control environment}
The model architecture for the continuous control environment is the same as the grid world, except that the input encoder uses MLPs of 2 hidden layers of 256 units each with ReLU activations, instead of CNNs.

\subsection{Sub-task policies}
\label{appendix:architecture}
The sub-task policies $\pi_\theta(a|s,z)$ are composed of a CNN module to embed the environment state $s_t$ and a subsequent MLP head to predict the probability of taking a particular action. This CNN shares the same architecture as the recognition model CNN. In initial experiments, we found that training separate policies $\pi_{\theta_z}(a|s)$ for each sub-task $z\in\{1,...,K\}$ with shared CNN parameters led to better generalization performance than embedding the sub-task latent variable and providing it as input to just a single policy for all sub-tasks. For continuously relaxed latent variables $z$, i.e.~during training, we use a soft mixture $\pi_\theta(a|s,z)=\sum_{k=1:K}q(z=k|a,s,b)\pi_{\theta_k}(a|s)$ to obtain gradients, where we have omitted time step and segment indices to simplify notation.

\subsection{Termination policy}
\label{appendix:termination}
To allow for our model to be used in an online setting where the end of an event segment has to be identified before ``seeing the future'', we jointly train a termination policy that shares the same model architecture (but without shared parameters) as the boundary prediction network $q_{\phi_b}(b_i|x)$, but with a $\mathrm{sigmoid}(x)=1/(1+e^{-x})$ activation function on the logits instead of a (Gumbel) softmax. It similarly passes over the input sequence $M$ times (with softly masked out RNN hidden states) and is trained to predict an output of $1$ (i.e., terminate) for the location of the $i$-th boundary $b_i = \mathrm{argmax}_{t={1:T}}q_{\phi_b}(b_{i}=t|x)$ and zero otherwise. At test time, we use a threshold of $0.5$ to determine termination.
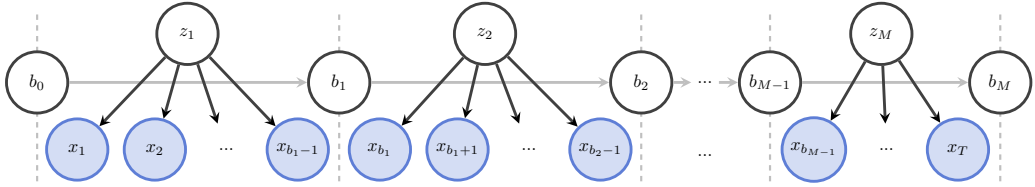
\begin{figure*}[htp]
\centering
\resizebox{0.8\linewidth}{!}{\begin{tikzpicture}
  \definecolor{mycolor}{RGB}{40,83,200}
  \tikzstyle{latent}=[circle,ultra thick,draw=black!75,fill=white,minimum size=12mm]
  \tikzstyle{observed}=[circle,ultra thick,draw=mycolor!75,fill=mycolor!20,minimum size=12mm]
  \tikzstyle{dots}=[minimum size=10mm]
  \tikzstyle{blackarrow}=[-stealth, ultra thick,draw=black!75]
  \tikzstyle{greyarrow}=[arrows={-stealth[black!25]}, ultra thick, draw=black!25]

    \node[] (0) {};
	\node[latent, right = 12.45em of 0] (15) {$b_1$};
	\node[above = 2em of 15] (16) {};
	\node[below = 4em of 15] (17) {};
	\node[latent, right = 13.25em of 15] (1) {$b_2$};
	\node[above = 2em of 1] (19) {};
	\node[below = 4em of 1] (20) {};
	\node[right = 1em of 1] (14) {$...$};
	\node[below = 3.25em of 14] (24) {$...$};
	\node[latent, right = 3.6em of 1] (21) {$b_{M-1}$};
	\node[above = 2em of 21] (22) {};
	\node[below = 4em of 21] (23) {};
	\node[latent, right = 9.25em of 21] (29) {$b_{M}$};
	\node[above = 2em of 29] (30) {};
	\node[below = 4em of 29] (31) {};
	\node[latent, left = 13.25em of 15] (32) {$b_{0}$};
	\node[above = 2em of 32] (33) {};
	\node[below = 4em of 32] (34) {};
	
	\node[above = 2em of 0] (10) {};
	\node[latent, right = 4em of 10] (11) {$z_1$};
	
	\node[above = 1em of 1] (12) {};
	\node[latent, right = 13em of 11] (13) {$z_2$};
	\node[latent, right = 18.5em of 13] (28) {$z_M$};

	\node[observed, below = 1.6em of 0] (3) {$x_1$};
	\node[observed, right = .75em of 3] (4) {$x_2$};
	\node[dots, right = .75em of 4] (5) {$...$};
	\node[observed, right = .75em of 5] (6) {$x_{b_{1}-1}$};
	
	\node[observed,right = 1em of 6] (7) {$x_{b_{1}}$};
	\node[observed, right = .75em of 7] (8) {$x_{b_{1}+1}$};
	\node[dots, right = .75em of 8] (9) {$...$};
	\node[observed, right = .75em of 9] (18) {$x_{b_{2}-1}$};
	
	\node[observed,right = 8.25em of 18] (25) {$x_{b_{M-1}}$};
	\node[dots, right = .75em of 25] (26) {$...$};
	\node[observed, right = .75em of 26] (27) {$x_T$};
	
	\path[greyarrow] (15) edge (1);
	\path[greyarrow] (1) edge (14);
	\path[greyarrow] (14) edge (21);
	\path[greyarrow] (21) edge (29);
	\path[greyarrow] (32) edge (15);
	
	\begin{scope}[on background layer]
	    \path[dashed, very thick, draw=black!25] (16) edge (17);
	    \path[dashed, very thick, draw=black!25] (19) edge (20);
	    \path[dashed, very thick, draw=black!25] (22) edge (23);
	    \path[dashed, very thick, draw=black!25] (30) edge (31);
	    \path[dashed, very thick, draw=black!25] (33) edge (34);
    \end{scope}
    
	\path[blackarrow] (11) edge (3);
	\path[blackarrow] (11) edge (4);
	\path[blackarrow] (11) edge (5);
	\path[blackarrow] (11) edge (6);
	
	\path[blackarrow] (13) edge (7);
	\path[blackarrow] (13) edge (8);
	\path[blackarrow] (13) edge (9);
	\path[blackarrow] (13) edge (18);
	
	\path[blackarrow] (28) edge (25);
	\path[blackarrow] (28) edge (26);
	\path[blackarrow] (28) edge (27);
\end{tikzpicture}}
\caption{Dependencies between observed and latent variables in our generative model $p_\theta(x_{1:T}|b_{1:M},z_{1:M})$. The state-action pair $(a_t, s_t)$ is summarized into a single observed variable $x_t$. The latent variables $b_i$ determine the location of the boundaries between segments, whereas $z_i$ summarize the content of each segment.}\label{fig:graphical_model} 
\end{figure*}
\subsection{ELBO objective for learning}
\label{appendix:elbo}
We jointly optimize for both the parameters of the sub-task policy $\pi_\theta(a|s,z)$ and the recognition model $q_\phi(b,z|a,s)$ by using the ELBO as an objective for learning:
\begin{align}
\ELBO &= \expt_{q_\phi(b, z|a,s)}[\log p_\theta(a|s,b,z) \nonumber\\&+ \log p(b, z) - \log q_\phi(b, z|a,s)],
\label{eq:elbo}
\end{align}
where we have dropped time step and sub-task indices for ease of notation. The first term can be understood as the (negative) reconstruction error of the action sequence, given a sequence of states and inferred latent variables, whereas the last two terms, in expectation, form the Kullback-Leibler (KL) divergence between the prior $p(b, z)$ and the posterior $q_\phi(b, z|a,s)$. The ELBO can be obtained from the original BC objective as follows, using Jensen's inequality:
\begin{align}
\log p_\theta(a|s) &= \log\sum_{b,z}p_\theta(a|s,b,z)p(b,z) \nonumber\\
&= \log\,\expt_{q_\phi(b, z|a,s)}\left[\frac{p_\theta(a|s,b,z)p(b,z)}{q_\phi(b, z|a,s)}\right] \nonumber\\
&\geq \expt_{q_\phi(b, z|a,s)}\left[\log\frac{p_\theta(a|s,b,z)p(b,z)}{q_\phi(b, z|a,s)}\right] \nonumber\\
&= \ELBO
\label{eq:elbo-deriv}
\end{align}

An overview of the dependencies between observed and latent variables in our model is provided in Figure \ref{fig:graphical_model}.

\subsection{KL term}
\label{appendix:kl-term}
We use a scale hyperparameter $\beta\in[0,1]$ to scale the contribution of the KL term in Eq.~\eqref{eq:elbo} similar to the $\beta$-VAE framework \cite{higgins2016beta}, which gives us control over the strength of the prior $p(b,z)$. As is common in applications of relaxed categorical posteriors in a VAE \cite{jang2016categorical}, we choose a simple (non-relaxed) categorical KL term for both the posterior distributions $q_{\phi_b}(b_i|x)$ and $q_{\phi_z}(z_i|x)$.

Further, as we do not know the precise location of the boundary latent variables $b_i$ at training time, we cannot evaluate $p(b_{i}|b_{i-1})$ for $i>1$ in the relaxed/continuous case. Under the assumption of independence between segments, behavior within each segment originating from the same distribution, and with a shared recognition model for all latents, see Eq.~\eqref{eq:recognition-model}, we can equivalently evaluate the KL term related to $b$ for the first boundary only, i.e.~for $p(b_{1})$, and multiply this term by $M$, where $M$ is the number of segments (we use this setting in our experiments). Alternatively, one could place a prior on $\sum_{t=1:T}P(t\in C_i)$, which can be understood as a continuous relaxation of the length of a segment. This would allow for an individual KL contribution for every segment, which could be useful for other applications or environments, where our assumptions are too restrictive.

\subsection{Gaussian latent variables}
\begin{figure*}[ht!]
    \quad\begin{subfigure}[b]{0.39\textwidth}
        \centering
        \includegraphics[width=\textwidth]{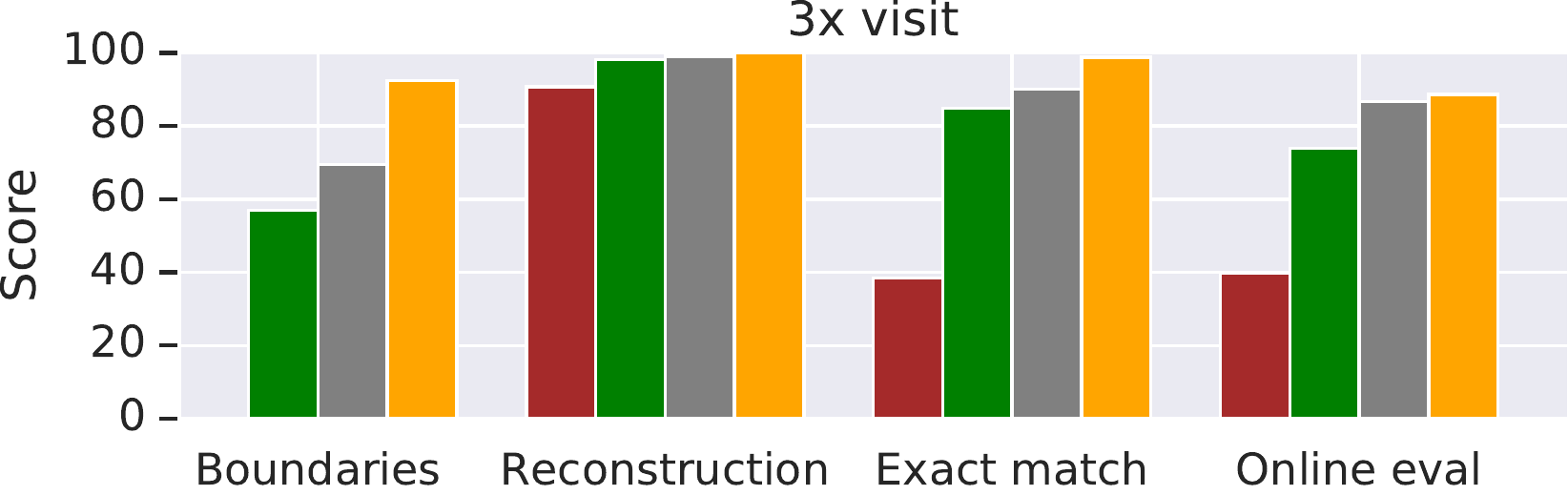}
    \end{subfigure}%
    ~\quad
    \begin{subfigure}[b]{0.5\textwidth}
        \centering
        \includegraphics[width=\textwidth]{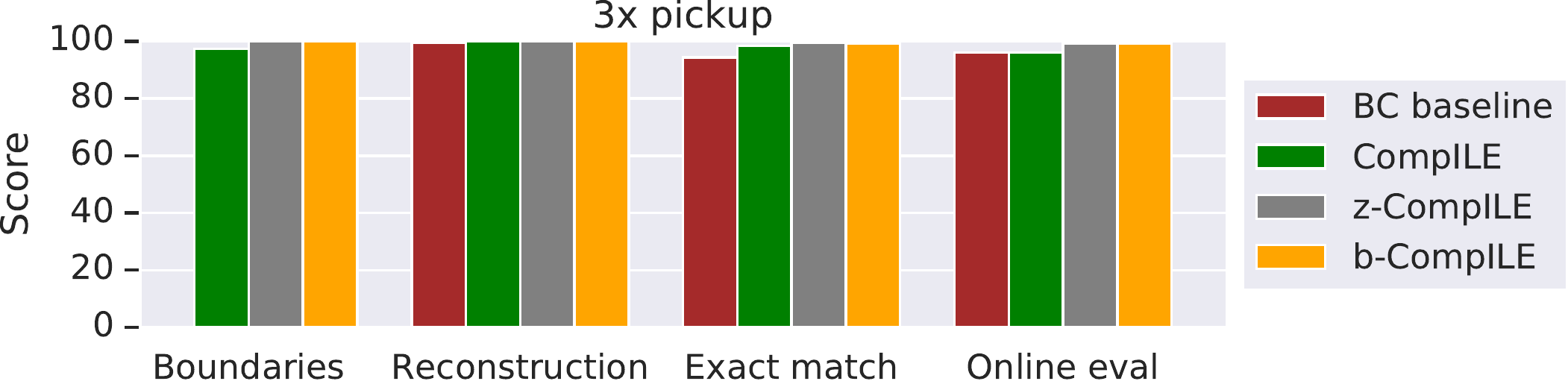}
    \end{subfigure}%
    \\\vspace{0.1em}
    \,\,\,\,\begin{subfigure}[b]{0.39\textwidth}
        \centering\vspace{0.5em}
        \includegraphics[width=\textwidth]{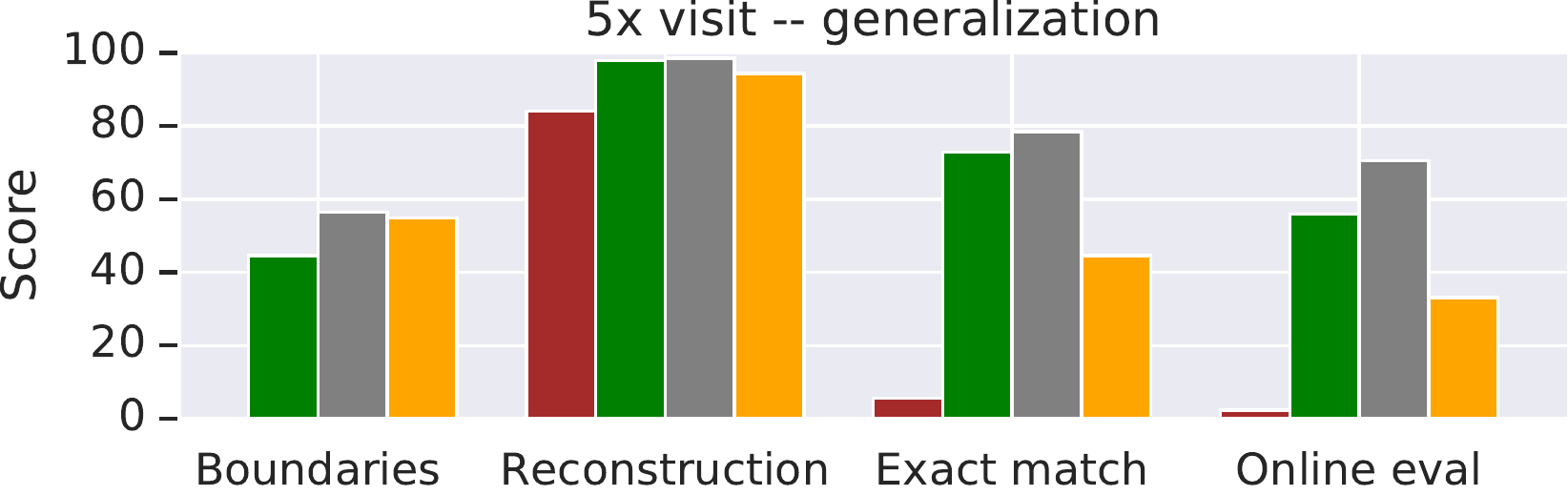}
    \end{subfigure}%
    ~\quad
    \begin{subfigure}[b]{0.39\textwidth}
        \centering\vspace{0.5em}
        \includegraphics[width=\textwidth]{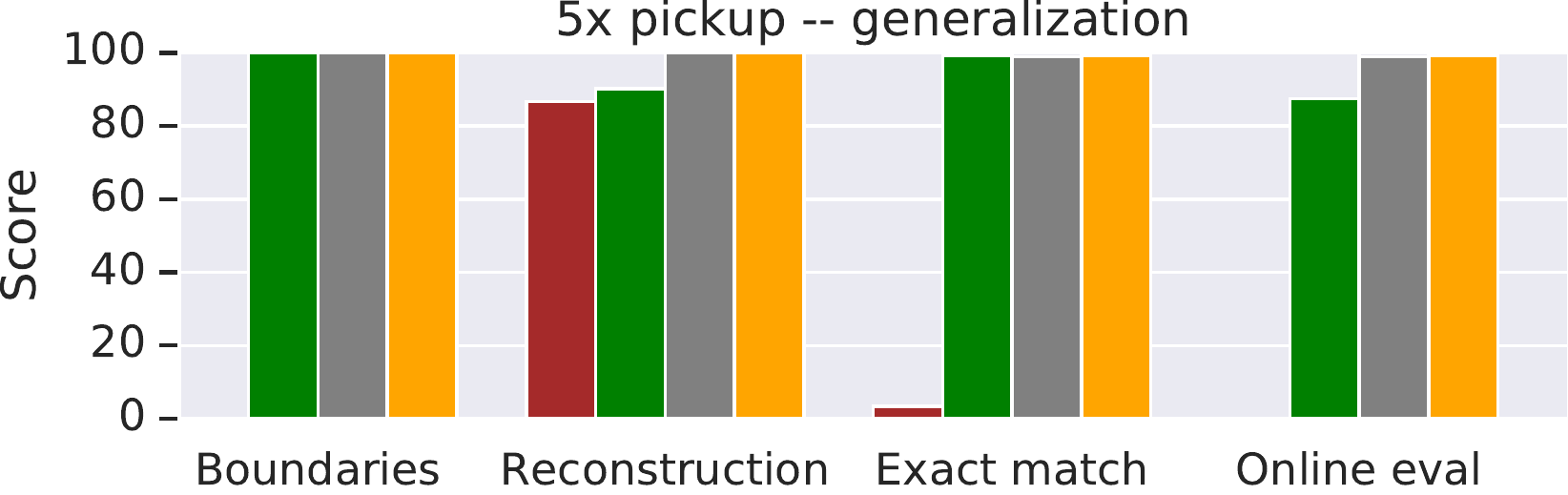}
    \end{subfigure}%
    
    {\centering
    \caption{Imitation learning results in grid world domain for CompILE model variant with Gaussian latent variables.\label{fig:imitation_results_gauss}}} 
\end{figure*}
We experimented with continuous, Gaussian latent variables $z$ in the grid world domain and found that our model can support this setting with only minor modifications. We use a single policy $\pi_\theta(a|s,z)$ for decoding, where the MLP head takes the latent variable $z$ (passed through a single, trainable linear layer) as input in addition to the state embedding (both are concatenated). We further place a unit-variance, zero-mean Gaussian prior on $z$ and use the appropriate KL term. We trained and tested this model variant under the same setting as the experiments with discrete latent variables, with the exception of using 32-dimensional Gaussian latent variables. Results for this setting are summarized in Figure \ref{fig:imitation_results_gauss} for the grid world domain and in Table \ref{tab:cont_control_results_gaussian} for the continuous control domain. 

\begin{table}[t]
\centering
\resizebox{0.96\columnwidth}{!}{%
\begin{tabular}{rccc}
\toprule
\textbf{Model} & \textbf{Accuracy} & \textbf{F1 (tol=0)} & \textbf{F1 (tol=1)}  \\
\midrule
\multicolumn{4}{c}{3 tasks} \\
LSTM surprisal & $ 24.8 \pm 0.6 $ & $ 39.0 \pm 0.3 $ & $ 47.1 \pm 0.4 $  \\
CompILE & $ 45.2 \pm 13.8$ & $ 59.3 \pm12.1$ & $68.8 \pm 8.1 $ \\
z-CompILE & $ 99.6 \pm 0.2 $ & $ 99.6 \pm 0.1 $ & $ 99.9 \pm 0.1 $ \\
b-CompILE & $ 99.8 \pm 0.1 $ & $ 99.9 \pm 0.1 $ & $ 99.9 \pm 0.0 $ \\
\midrule
\multicolumn{4}{c}{5 tasks -- generalization} \\
LSTM surprisal & $ 21.6 \pm 0.5 $ & $ 44.9 \pm 0.5 $ & $ 54.4 \pm 0.5 $ \\
CompILE & $ 28.7 \pm 7.0 $ & $ 56.4 \pm 9.1 $ & $ 63.8 \pm 6.5 $ \\
z-CompILE & $ 98.3 \pm 0.5 $ & $ 99.2 \pm 0.3 $ & $ 99.7 \pm 0.1 $ \\
b-CompILE & $ 98.5 \pm 0.3 $ & $ 99.3 \pm 0.2 $ & $ 99.7 \pm 0.1 $ \\
\bottomrule
\end{tabular}}
\caption{Segmentation results in continuous control domain for CompILE model variant with Gaussian latent variables. Values are in \% and we report mean and standard deviation for runs with 5 different random seeds.\label{tab:cont_control_results_gaussian}}
\vspace{-1em}
\end{table}

\begin{table}[t]
\centering
\resizebox{0.9\columnwidth}{!}{%
\begin{tabular}{rccc}
\toprule
\textbf{Hyperparam.} & \textbf{Accuracy} & \textbf{F1 (tol=0)} & \textbf{F1 (tol=1)}  \\
\midrule
\multicolumn{4}{c}{\# Segments $M$} \\
$M=3$ & $ 62.0 \pm 4.5 $ & $ 74.3 \pm 3.3 $ & $ 78.9 \pm 2.5 $  \\
$M=4$ & $ 53.4 \pm 6.3 $ & $ 66.6 \pm 5.5 $ & $ 75.3 \pm 3.1 $ \\
$M=5$ & $ 29.8 \pm 6.1 $ & $ 47.3 \pm 5.5 $ & $ 65.6 \pm 2.9 $ \\
\midrule
\multicolumn{4}{c}{Softmax temperature $\tau$} \\
$\tau=1$ & $ 62.0 \pm 4.5 $ & $ 74.3 \pm 3.3 $ & $ 78.9 \pm 2.5 $   \\
$\tau=0.1$ & $ 40.9 \pm 5.3 $ & $ 55.3 \pm 5.7 $ & $ 67.0 \pm 2.4 $  \\
$\tau=0.01$ & $ 36.3 \pm 6.0 $ & $ 50.3 \pm 5.2 $ & $ 66.4 \pm 3.6 $ \\
\bottomrule
\end{tabular}}
\caption{Segmentation results in continuous control domain with 3 tasks for CompILE model trained with (a) different number of segments ranging from $M=3$ (correct setting) to $M=5$ (too many boundaries provided to the model), and (b) softmax temperature ranging from $\tau=1$ to $\tau=0.01$. Values are in \% and we report mean and standard deviation for runs with 5 different random seeds.\label{tab:cont_control_results_segments}}
\vspace{-1em}
\end{table}

\subsection{Attentive readout}
Instead of (softly) reading the logits for the latent variables $z_i$ from the last time step within a segment, we experimented with using a learned attention mechanism, masked by the respective soft segment mask. In this setting, we add another output head (a single, learnable linear layer) on top of the recognition model RNN which we denote by $a_i^t$, where $t$ stands for the time step and $i$ denotes the segment index. Before passing the attention scores $a_i^t$ through a softmax layer, we re-normalize using the segment probability $P(t\in C_i)$:
\begin{equation}
\tilde{a}_i^t = a_i^t + \log P(t\in C_i),
\end{equation}
i.e.~we softly mask the attention scores so that the read-out is only performed within the respective segment. The final attention score is obtained as $s_i = \mathrm{softmax}(\tilde{a}_i)$, where the softmax is applied over the time dimension. We read out the logits of $z_i$ from the output heads as follows:
\begin{align}
\label{eq:output_head_masking_attn}
q_{\phi_z}(z_i|x) = \mathrm{concrete}_\tau(\sum_{t=1:T}s_i^t h_{z_i}^t).
\end{align}
We found that results were similar in both settings and that the model typically learned to attend to the last time step within the segment. For different environments where the cue for a specific sub-goal in a segment of behavior appears at different locations within the segment, the attention mechanism will potentially be a better fit than a soft read-out at the end of the segment.

\subsection{Other hyperparameters}

\paragraph{Number of hidden units and MLP layers} We use 256 hidden units in all MLP layers and in the LSTM throughout all experiments, unless otherwise mentioned. A smaller number of hidden units mostly did not affect the boundary prediction accuracy, but slightly reduced performance in terms of reconstruction accuracy. For the output heads for $h_{z_i}$, we use a single, trainable linear layer (we experimented with deeper MLPs but didn't find a difference in performance) and we use a single hidden layer MLP with ReLU activation function for the output head $h_{b_i}$ (the output is a scalar for every time step). Similarly, the policy MLP is using a single hidden layer with ReLU activation in the maze task, while for the control task we used a 2 layer MLP. The termination policy uses an MLP with two hidden layers with ReLU activation functions on top of the RNN outputs.

\paragraph{Number of segments} 
The hyperparameter $M$, i.e., the number of segments that the model is allowed to use to explain a particular input sequence, can have an impact on reconstruction and segmentation quality. We generally find that we obtain best results by providing the model with the true number of underlying segments (if this number is known). When providing the model with more than necessary segments, it often learns to place unneeded segmentation boundary indicators at the end of the sequence, while in some cases the model over-segments the trajectory (i.e., it breaks a single segment into parts). We provide results for this setting on the continuous control task in Table \ref{tab:cont_control_results_segments}, and we find that the accuracy (and F1 score) for segmentation boundary placement slightly degrades if the model is provided with more than necessary segments.

\paragraph{Poisson prior rate} We fix the Poisson rate to $\lambda=3$ in all experiments. We found that our model was not very sensitive to the precise value of $\lambda$.

\paragraph{Softmax temperature} We experimented with annealing the Gumbel softmax temperature over the course of training, starting from a temperature of 1 and found that it could slightly improve results, depending on the precise choice of annealing schedule and final temperature. To simplify the exposition and to allow for easier reproduction, however, we report results with fixed temperature of 1 throughout training unless otherwise mentioned. In Table \ref{tab:cont_control_results_segments}, we provide results for experiments with lower softmax temperature (fixed throughout training) on the continuous control task. We found that the boundary prediction accuracy degrades when training with lower temperatures without annealing. When training with partial supervision on either the boundary positions (b-CompILE) or segment encodings (z-CompILE), we found that results are unaffected by lower softmax temperatures.

\section{Reinforcement learning agent details}
\label{appendix:rl}
\subsection{Architecture and hyperparameters}
The agent uses a smaller model than our CompILE imitation learning model, but otherwise similarly has a 2-layer CNN encoder followed by an MLP policy. The CNN has $3\times 3$ filters with 32 feature maps, followed by an MLP with two hidden layers of size 128. Both the CNN and the MLP use ReLU activations. All agents use the same architecture, and the hierarchical agent based on the pre-trained CompILE model uses 128 instead of 256 hidden units (otherwise same training and same architecture as in the imitation learning experiments). The hierarchical agent has access to both low-level actions (8 in total) and 10 meta-actions which correspond to executing one sub-policy of the CompILE model.

The baseline VAE-based BC agent corresponds to an ablation of the hierachical CompILE-based agent, where we use only a single segment (i.e.~$M=1$, no segmentation) during training and a 128-dimensional categorical latent variable $z$ (instead of 10 categories). The agent therefore can choose between 128 meta-actions and 10 low-level actions.

We embed the current task type (visit or pick up) and object type each in a 16-dim vector, via a trainable linear layer. These are concatenated and provided to the policy model in the following two ways: 1) we concatenate this embedding vector with the current observation along the channel (object type) dimension before we feed it into the CNN, and 2) we concatenate the embedding vector with the last hidden layer of the policy MLP. The former allows the CNN to be conditioned on the task type, while we found the second concatenation in the policy MLP to help convergence. For the VAE-based BC baseline (which tries to solve multiple tasks at once), we do not just provide the current task, but the full list of remaining tasks by embedding each task and concatenating them into a single vector (with zero-padding for already fulfilled tasks).

For IMPALA \cite{espeholt2018impala}, we use an entropy cost factor of $0.0005$, a baseline cost factor of $0.5$, and a discounting factor of $0.99$. The agents are trained with the Adam optimizer \cite{kingma2014adam} using a learning rate of $0.001$ and a batch size of $256$.

\subsection{Distributed training}
We distribute the training of this agent into one learner and multiple actors following the IMPALA framework \cite{espeholt2018impala}, where the actors generate trajectories using the current agent parameters for training, and the learner updates the agent parameters based on the trajectories received from the actors.  The learner runs on a GPU, while the actors run on CPUs.  The number of actors is tuned to maximize the throughput of the learner.

This framework uses the actor-critic training algorithm, with off-policy correction \cite{espeholt2018impala} to handle the staleness of the actor generated trajectories.  This correction is necessary as the actors and the learner are not always in sync in a distributed setting, and the parameter weights used for generating trajectories are usually not the latest learner weights when the learner receives the trajectories.

\section{Environment implementation details}
\label{appendix:environment}
\subsection{Grid world} The environment is implemented in pycolab (\url{https://github.com/deepmind/pycolab}) with 8 different primitive actions: move north, move east, move south, move west, pick up north, pick up east, pick up south, pick up west. Each executed action corresponds to one time step in the environment. Observations $s_t$ are tensors of shape $10\times10\times N_{\mathrm{things}}$, where $N_{\mathrm{things}}$ is the total number of things available in the environment, in our case these are 10 object types that can be interacted with, impassable walls and the player, i.e.~$N_{\mathrm{things}}=12$. We ensure that the task is solvable and no walls make objects unreachable. Walls are placed using a recursive backtracking algorithm for unbiased maze generation. We further subsample walls using a sampling rate of 0.2 to simplify the task. The 2D grid is enclosed by a single row/column of walls that are not subsampled.

Demonstration sequences are generated using a breadth-first search on the graph defined by all allowed movement transitions to find the shortest path to the goal object (ties are broken in a consistent manner). For pick up instructions, we replace the last move action in the demonstration sequence with a directional pick up action. We cut demonstration sequences to a maximum length of 42 at training time, and 200 at test time (as some of our tests involve more tasks).

\subsection{Continuous control} 

This environment is adapted from the single target reacher task in DeepMind control suite \cite{tassa2018deepmind}. The reacher arm is composed of two segments, each with length $l=0.12$, and the controller controls the two motors on the two joints of the arm, one at the shoulder and the other at the elbow.  The control actions are the angular velocities to be applied at the two joints.  Target objects (spheres) have a diameter of $d=0.05$, and they are placed in a belt around the center, with the distance to the center sampled uniformly from range $[0.05, 0.2]$, and direction (angle) sampled uniformly around the circle.  The environment is set up to take control actions in time intervals of $0.06$, with each episode taking a maximum time of 6, i.e.~$100$ time steps at most.

In this customized environment, we have a total of $K=10$ distinct target types, each designated with a different color in the rendered scenes.  Each target is represented using 3 numbers ($\alpha$, $x$, $y$), where $\alpha$ is the visibility of the object, and $\alpha=1$ if the object is visible, and $\alpha=0$ otherwise, $(x, y)$ is the Cartesian coordinate of the target.

In each episode, we first set the number of tasks to $M=3$ or $M=5$, and then sample the number of objects $N$ in range $[M, 6]$ uniformly, and then pick $M$ out of $N$ objects uniformly without replacement as the targets to create a task list.

The agent receives an observation that is composed of 2 parts, the first part is the concatenation of all object tuples, arranged in a vector like this: $(\alpha_1, x_1, y_1, \alpha_2, x_2, y_2, ..., \alpha_K, x_K, y_K)$, where $(\alpha_i, x_i, y_i)$ describes the $i$th object type.  If the $i$th object type is not selected (not among the $N$ objects being selected) in this episode, then all of $\alpha_i$, $x_i$ and $y_i$ are set to 0.  The second part is the position of the reacher arm represented as two angles $(\theta_1, \theta_2)$, where $\theta_1$ is the angle at the shoulder joint, and $\theta_2$ is the angle at the elbow joint.

The coordinate of the finger tip of the arm is computed as $(l\cos\theta_1 + l\cos(\theta_1 + \theta_2), l\sin\theta_1 + l\sin(\theta_1 + \theta_2))$.  A target is considered reached if this coordinate is within the sphere for the given target.  

Once a target is reached, the $\alpha$ value for that target is set to 0 (but the $x$ and $y$ values remain in the observation), and in the next time step the environment advances to the next task, with a new target being selected as the goal.

The demonstration trajectories are generated by a hand-designed controller.  The controller has access to the coordinates of the next target.  It first computes the coordinates of the finger tip, and then computes (1) the distance of the finger tip to the center (where the shoulder joint is); and (2) the angle of the finger tip.  If the distance is smaller than the distance of the target to the center, the elbow motor applies an angular velocity to open the arm (so that the finger tip can reach further), and if the distance is larger then the elbow closes.  On the other hand if the direction of the arm does not align with the target, the shoulder motor then applies an angular velocity to rotate the arm toward the target.

\section{Evaluation details}
\label{appendix:evaluation}
\subsection{Metrics}
In the imitation learning experiments in Section \ref{sec:imitation-learning}, we report the following four evaluation metrics:
\begin{itemize}
    \item \textbf{Boundaries}: We measure the accuracy of predicted boundary position. For each boundary latent variable $b_i$, we check if it exactly matches the ground truth task boundary, i.e., the point where a task ends and a new task begins. Let $b_i$ denote the ground truth position for the $i$-th boundary, then the accuracy is defined as
    $$
    \frac{1}{M-1}\sum_{i=1}^{M-1} \mathbb{I}[\argmax_{b_i} q_{\phi_b}(b_i|x) = b_i],
    $$
    where $\mathbb{I}[x=y]$ denotes the Iverson bracket that returns $1$ if $x=y$ and $0$ otherwise.
    \item \textbf{Reconstruction}: This measures the average reconstruction accuracy of the original action sequence, given the ground truth state sequence, i.e., in a setting similar to teacher forcing:
    $$
    \frac{1}{T}\sum_{i=1:M}\left(\quad\,\,\sum_{\mathclap{j=b_{i'}:b_{i}-1}}\,\,\mathbb{I}[\argmax_{a_j} \pi_{\theta}(a_j|s_j,z_i) = a_j]\right) ,
    $$
    where $i'=i-1$ and $b_i = \argmax_{b_i} q_{\phi_b}(b_i|x)$.
    \item \textbf{Exact match}: Here we measure the percentage of \textit{exact matches} of full reconstructed action sequence (i.e., this score is 1 if all actions match for a single demonstration sequence and 0 otherwise), given the ground truth state sequence (provided one step at a time) as input.
    \item \textbf{Online eval}: Here, we first run our recognition model on a demonstration trajectory to obtain a sequence of latent codes. Then, we run the sub-task policy corresponding to the first latent code in the environment, until the termination policy predicts termination, in which case we move on to the next latent code, run the respective sub-task policy, and so on. We terminate if the episode ends (more than 200 steps, wrong object picked up or all tasks completed) and measure the obtained reward (either 0 or 1). For the baseline model, we infer a single latent code and run the respective policy until the end of the episode (without termination policy). We report the average reward obtained (multiplied by a factor of 100).
    \item \textbf{F1 Score}:  To evaluate the pointer prediction performance for the continuous control task, we use the extra metric $F1$ score and optionally with a tolerance.  In the continuous control setting, it is not easy to get the boundaries exactly correct as the transitions of the actions and observations across time steps are mostly smooth.  The $F1$ score treats the predicted pointer locations and the ground truth pointer locations as 2 sets, and compute the precision as 
    $$
    \frac{\text{\#predictions that matches the ground truth}}{\text{total \#predictions}},
    $$
    irrespective of ordering, and recall as
    $$
    \frac{\text{\#ground truth that has matches in predictions}}{\text{total \#ground truth}}.
    $$
    The $F1$ score is computed as 
    $$
    F1 = \frac{2 \cdot \mathrm{precision} \cdot \mathrm{recall}}{\mathrm{precision} + \mathrm{recall}}.
    $$
    A `match' is considered to be successful if a predicted pointer location exactly equals a ground truth pointer location.  With tolerance 1, a match is considered successful if the two are off by at most 1 time steps.
\end{itemize}

\subsection{Segmentation baseline (LSTM surprisal)}

To compare segmentation performance, we implemented a baseline algorithm based on auto-regressive behavioral cloning, termed \emph{LSTM surprisal}. Given the state-action sequence $((s_1, a_1), (s_2, a_2), \ldots, (s_T, a_T))$, this model maximizes the likelihood in the following form:
\begin{equation}
    \max_{\theta} P_{\theta}(a_{1:T} | s_{1:T}) = \prod_{i=1}^T P(a_i | a_{1:i-1}, s_{1:i})
\end{equation}

Then, a natural approach to decide the segment boundary is based on the probability of each action. An action which is surprising (i.e., having low conditional probability) to the model should be an action that marks the beginning or end of a task segment.

Given the number of chunks $M$, we find the top $M-1$ boundary indicator variables $b_1, b_2, \ldots, b_{M-1}$ with minimum conditional likelihood, i.e.,
\begin{equation}
     \argmin_{[b_1, b_2, \ldots, b_{M-1}], b_i \leq b_{i+1}} \sum_{i=1}^{M-1} P(a_{b_i} | a_{1:b_i-1}, s_{1:b_i})
\end{equation}
In the experiments, we use the same CNN (MLP for continuous control) architecture for encoding the state as in CompILE. An LSTM with same embedding size as our CompILE model is used here to model the dependency on the history of states and actions. We use the same training procedure as in the other models, i.e., we only train on 3x tasks, but report performance both on 5x. Interestingly, this model finds boundaries more consistently in the generalization setting (5 tasks) for the pick up task than in the setting it was trained on (3 tasks) in the grid world domain. We hypothesize that this is due to the fact that it has never seen a 4-th and 5-th object being picked up during training, and therefore assigns low probability to these events, which corresponds to a large ``surprise'' when these are observed in the generalization setting.

\section{Qualitative results}
\label{appendix:qualitative}
\begin{figure*}[htp]
  \centering
    \includegraphics[width=0.8\textwidth]{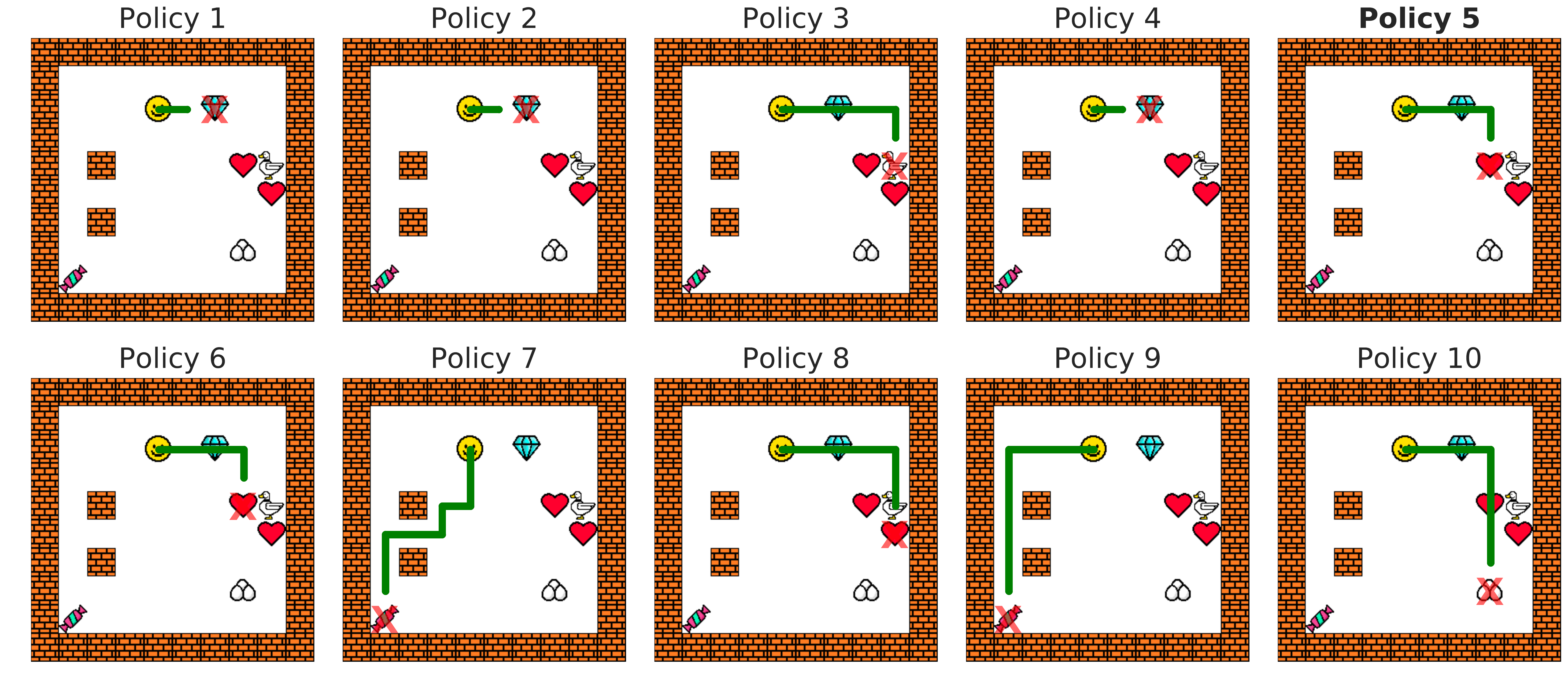}
  \caption{Example of sub-task policies discovered by the agent. \label{fig:policies_4}}
\end{figure*}

\begin{figure*}[htp]
  \centering
    \includegraphics[width=0.8\textwidth]{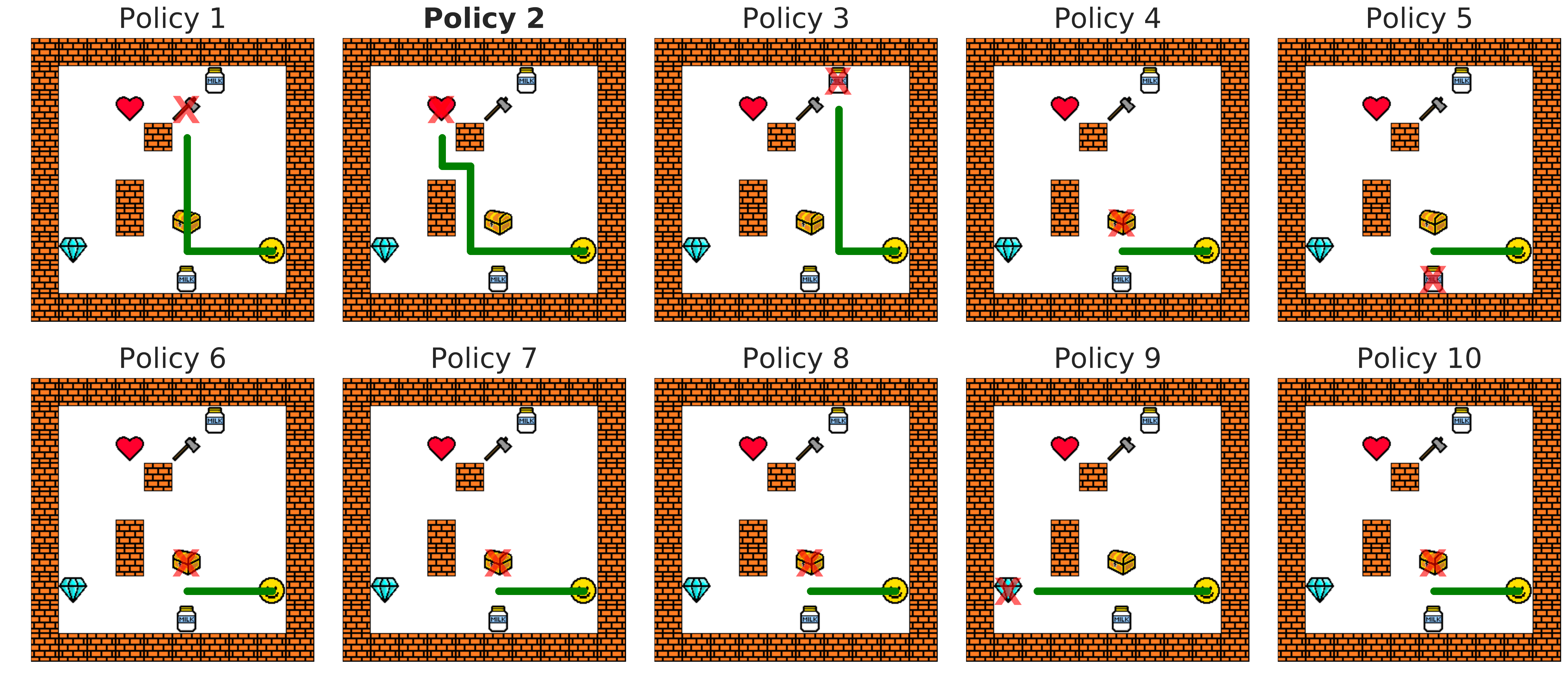}
  \caption{Example of sub-task policies discovered by the agent. \label{fig:policies_5}}
\end{figure*}

Here, we provide qualitative analysis of the discovered sub-task policies in the grid world environment. We run each sub-task policy for the pick up task on a random environment instance until termination, see Figures \ref{fig:policies_4}--\ref{fig:policies_5}. The red cross marks the picked up object. We mark the policy in bold that the inference model of CompILE has inferred from a demonstration sequence for the task \emph{pick up heart}.

\begin{figure*}[htp]
  \centering
    \includegraphics[width=0.76\textwidth]{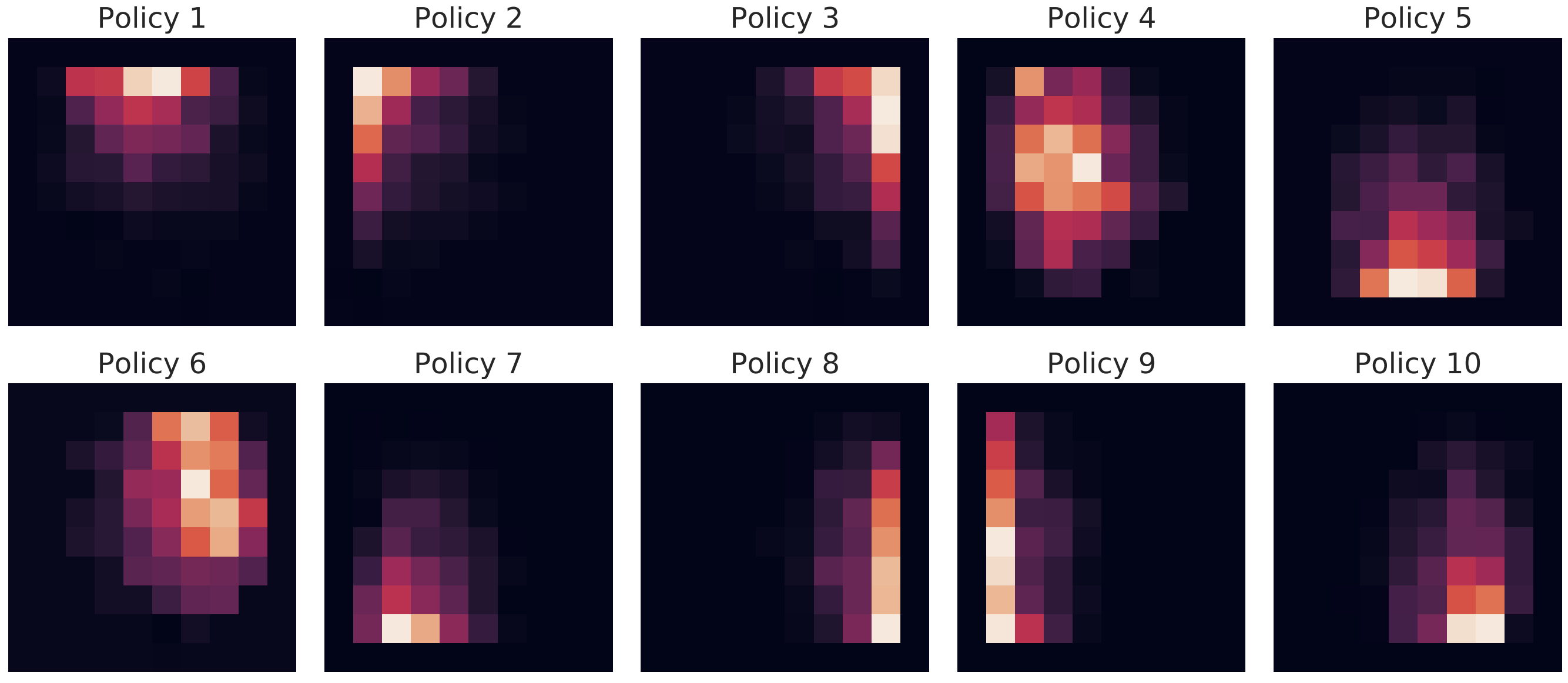}
  \caption{Heatmap of termination locations for each policy (for 1000 random environment instances). \label{fig:policies_heatmap}}
\end{figure*}

In Figure \ref{fig:policies_heatmap}, we investigate termination locations for the policies in the same trained CompILE model. We find that the model learns location-specific latent codes, which are effective at describing agent behavior from demonstrations. Nonetheless, the model can disambiguate close-by objects as can be seen in Figure \ref{fig:policies_4}.

\end{document}